\documentclass[journal]{IEEEtran}

\usepackage{times}
\usepackage{epsfig}
\usepackage{graphicx}
\usepackage{amsmath}
\usepackage{amssymb}
\usepackage[noend]{algpseudocode}
\usepackage{balance}
\usepackage{xcolor}
\usepackage{url}

\usepackage{multirow}

\newcommand{\etal}{\emph{et~al.}}

\usepackage{subfig}
\usepackage{cite}
\usepackage[linesnumbered,ruled]{algorithm2e}

\definecolor{black}{rgb}{0, 0, 0}

\graphicspath{ {figures/} }

\ifCLASSINFOpdf
\else
\fi

\begin{document}

\title{Elasticity Meets Continuous-Time:\\ Map-Centric Dense 3D LiDAR SLAM}

\author{Chanoh Park,~\IEEEmembership{Student Member,~IEEE,} Peyman Moghadam,~\IEEEmembership{Member,~IEEE,} Jason Williams,~\IEEEmembership{Member,~IEEE,} Soohwan  Kim,~\IEEEmembership{Member,~IEEE,} Sridha  Sridharan,~\IEEEmembership{Life Senior Member,~IEEE} and Clinton Fookes,~\IEEEmembership{Senior Member,~IEEE}       
\IEEEcompsocitemizethanks{\IEEEcompsocthanksitem $^1$ The authors are with the Robotics and Autonomous Systems Group, DATA61, CSIRO, Brisbane, QLD 4069, Australia.
E-mails: {\tt\small \emph{Chanoh.Park, Peyman.Moghadam, Jason.Williams}@data61.csiro.au}
\IEEEcompsocthanksitem $^{2}$ The authors are with the School of Electrical Engineering and Computer Science, Queensland University of Technology (QUT), Brisbane, Australia.
E-mails: {\tt\small \emph{chanoh.park, peyman.moghadam, s.sridharan, c.fookes}@qut.edu.au}
\IEEEcompsocthanksitem $^{3}$ Soohwan Kim is with Division of Smart Automotive Engineering, Sun Moon University, South Korea, E-mail: {\tt\small \emph{kimsoohwan}@gmail.com}}%
}

\markboth{IEEE Transactions on Robotics ,~Vol.~xx, No.~x, xx~2020}%
{Shell \MakeLowercase{\textit{et al.}}: Bare Demo of IEEEtran.cls for IEEE Journals}

\maketitle

\begin{abstract}
Map-centric SLAM utilizes elasticity as a means of loop closure. This approach  reduces the cost of loop closure while still provides large-scale fusion-based dense maps, when compared to the trajectory-centric SLAM approaches. In this paper, we present a novel framework for 3D LiDAR based map-centric SLAM. Having the advantages of a map-centric approach, our method exhibits new features to overcome the shortcomings of existing systems, associated with multi-modal sensor fusion and LiDAR motion distortion. This is accomplished through the use of a local Continuous-Time (CT) trajectory representation. Also, our surface resolution preservative matching algorithm and Wishart-based surfel fusion model enables non-redundant yet dense mapping. Furthermore, we present a robust metric loop closure model to make the approach stable regardless of where the loop closure occurs. Finally, we demonstrate our approach through both simulation and real data experiments using multiple sensor payload configurations and environments to illustrate its utility and robustness.

\end{abstract}


\IEEEpeerreviewmaketitle

\section{Introduction}\label{sec:introduction}

\begin{figure*}[t]
\centering{

\includegraphics[width=.78\textwidth]{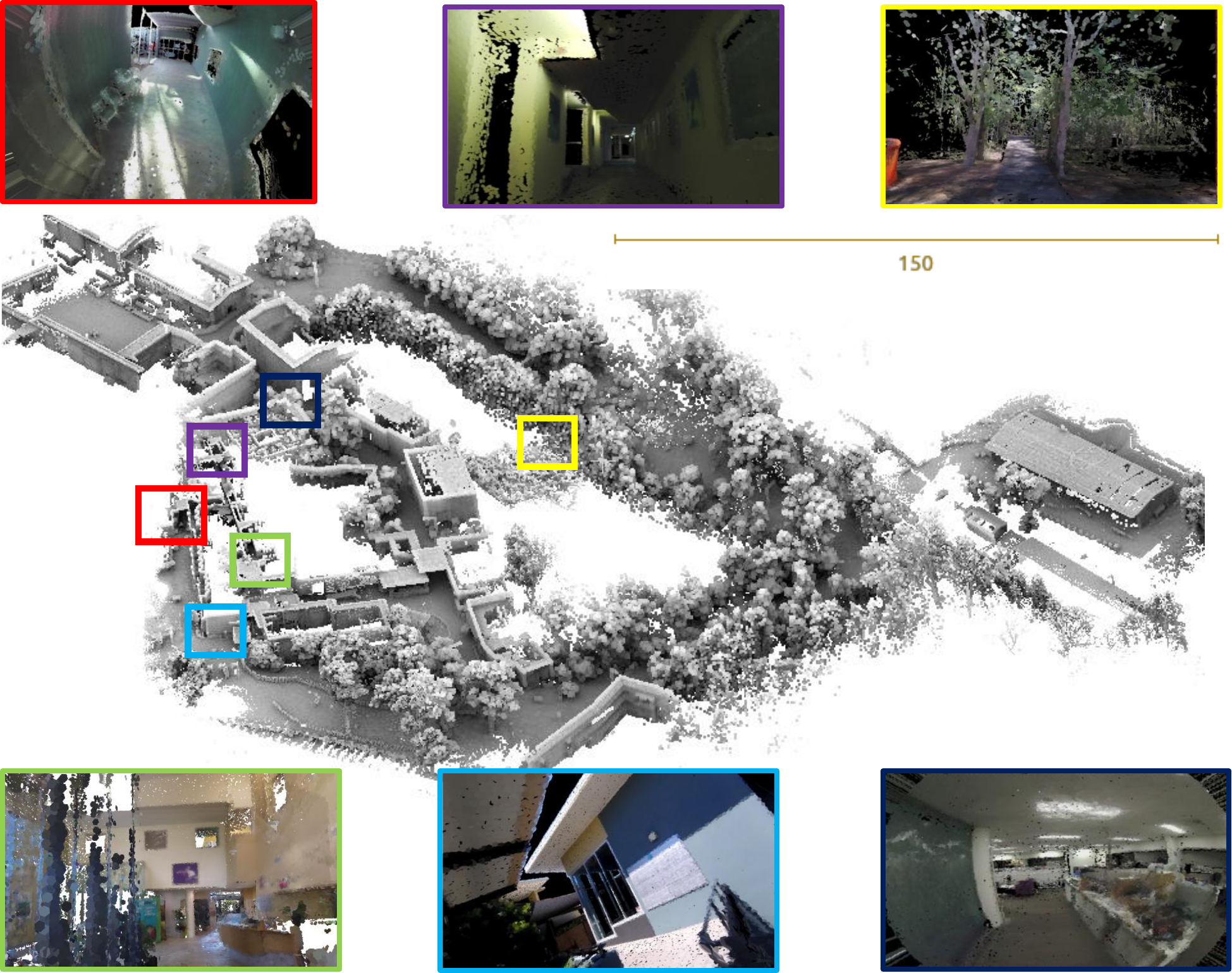}
}
\caption{
Map-Centric Dense 3D LiDAR SLAM: Reconstructed surfel map of a mixed environment with various sensors and platforms. The point cloud on the center is composed of seven datasets and shaded by PCV. Images shows the rendered colorized surfel cloud at different locations with various camera setup.
}

\label{fig:mainfiugre}       %
\end{figure*}

\IEEEPARstart{M}{ap}-centric SLAM solutions \cite{whelan2015, park2017c}, which have demonstrated their accuracy and effectiveness by fusion-based mapping and deformation-based loop closure, provide an alternative solution to the dominant trajectory-centric LiDAR SLAM \cite{bosse2012,vidas2015,shin2020}.
Due to the benefit that its complexity does not increase over time, the high quality reconstruction can be easily achieved by having redundant scans and fusing them.  
Despite their success, previous map-centric approaches have a particular set of drawbacks~\cite{whelan2015,behley2018rss,whelan2013,hess2016,park2017c}. Firstly, the original map-centric work is dedicated to a pinhole camera model such as RGB-D \cite{whelan2015} or multi-beam LiDARs \cite{behley2018rss}. 
Secondly, asynchronous sensor fusion and handling high motion distortion of LiDAR ~\cite{furgale2012} are not well studied within the map-centric framework. This is far from ideal as LiDAR-based systems often need to be tightly fused with multi-modal sensors to handle the motion distortion effects. Lastly, the map-centric approach is extremely susceptible to loop closure failure. The sensor data history cannot be stored due to the extensive input data size which implies that, once incorrectly fused, the map is not recoverable. 

In this paper, we seek an algorithm that exhibits the desirable characteristics of the map-centric approach, but is able to handle these problematic situations described above. We propose to do this by introducing the Continuous-Time framework into the map-centric approach. %
 The detailed contributions of this paper are as follows:
\begin{itemize}
  \item We introduce a new constraint for local Continuous-Time trajectory optimization to operate like a map-centric approach.
  \item We detail the realization of a new Continuous-Time trajectory optimization approach which is dedicated to the map-centric approach, improving local trajectory estimation accuracy.
  \item We derive a novel probabilistic surfel representation using the Wishart model. To the best of our knowledge, this is the first approach to introduce the Wishart model for surfel fusion. 
  \item We introduce a surface resolution preservative surfel matching method for non-pinhole type sensor data.
  \item We implement a LiDAR only sequential metric localization to reduce failure on loop closure.
  \item We demonstrate the solution on various environments and sensor configurations using simulation and real datasets.
\end{itemize}

Our previous work in \cite{park2017b,park2017c,park2019} presented an approach for LiDAR mapping using probabilistic surfel fusion, a method for map deformation, and a method for robust Visual-LiDAR metric localization respectively. These approaches deal with the issues separately and independently. In this work, we reformulate the map-centric 3D dense mapping problem with our new trajectory optimization method and intensively revise the dense surfel fusion model. Furthermore, in the metric localization stage we weaken the dependency of the vision to reduce the uncertainty from unobservable parameters related to the vision while maintaining the accuracy. Finally, we evaluate, analyze, and demonstrate this new approach in a comprehensive manner across various sensing platforms and across different environments (Figure \ref{fig:mainfiugre}) using simulation and real datasets to reveal its utility and robustness.

\section{Related Work}
\subsection{Map-centric Approach}

ElasticFusion \cite{whelan2015} proposed a {map-centric approach} for RGB-D cameras that removes pose graph optimization yet performs globally consistent mapping by giving elastic property to the map and directly deforming the map. The concept of map-centric SLAM eliminated the need of a pose graph for globally consistent mapping and converted the time dependency of the global optimization to {a} space-dependent problem. Also, by confining the tracking and fusion within recent map elements, they drastically reduced the processing time per input frame.

Despite these improvements in ElasticFusion method, {some features of their approach} are limited to RGB-D sensors and are not applicable to a LiDAR sensor model \cite{park2017b}.
Firstly, the original frame-to-frame motion model  \cite{whelan2015} is not an effective framework for handling asynchronous estimation and severe motion distortion of LiDAR.
Secondly, the map fusion method which is based on a pin-hole camera model can be applied only to the LiDAR configuration with low vertical field of view \cite{behley2018rss}. For example, the pin-hole camera model cannot be applied for the spinning LiDAR system \cite{bosse2012,park2017d} where the scan range often covers 360 degree both vertically and horizontally.  Thirdly, conventional LiDAR loop closure models \cite{latif2013} are not suitable for the map-centric approach. The second and third points will be elaborated in further detail in the following sections.

In a broad sense, sub-map dividing and realignment approaches \cite{droeschel20182,hess2016} are also similar to the map-centric method due to its property that input frames are fused into the local submaps. While the approaches in this category successfully reduced global optimization cost, the map fusion policy between graph nodes cannot be defined which brings the discontinuity issue in the map maintenance.

Recently, Behley \etal{}~\cite{behley2018rss} presented an approach analogous to the concept of the map-centric where the new estimations are fused into the global map. Although, the density of their map representation is dedicated for localization rather than dense mapping, their result present an important direction toward the fusible LiDAR map representation along with our previous research on LiDAR surfel fusion \cite{park2017b}.

\subsection{Map Fusion}

\begin{figure}[t]
 \centering
 \subfloat[Mesh, raw points, voxels, surfels]{\includegraphics[height=.09\textheight]{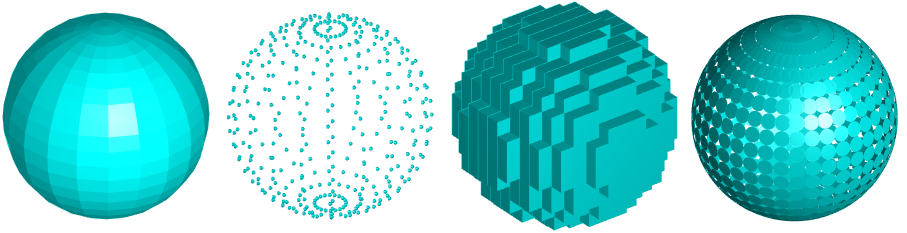}}\\
  \subfloat[Surfel rendering by shader]{\includegraphics[height=.13\textheight]{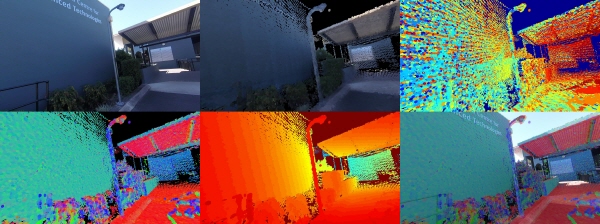}}
 \caption{(a) Comparison of different map representations: mesh, raw points, voxels and surfels. (b) Surfel map rendering example. From top left to right: image from camera, rendered colour surfel map, index map for colour fusion where each colour represents the identification number of a point. From bottom left to right:  normal map, depth map, blended colour and normal image.}
 \label{fig:maprepresentation}
 \end{figure}
 
\subsubsection{Map Representation and Fusion Feasibility}
The fusion between the global map and input frame is an essential step for high-quality map reconstruction and frame-to-model localization. A number of techniques exist for map representation and fusion in SLAM as shown in Figure \ref{fig:maprepresentation} (a). 

Most common dense map representation is using raw point cloud \cite{droeschel2018}. While it is most simple and efficient way to visualize the reconstruction result, the point representation is not proper for map element wise fusion and also requires relatively far more number of map elements to make a dense visualization. 
The approaches that digitize the space \cite{elfes1989,hornung2012,newcombe2011,whelan2012} successfully modeled probabilistic fusion of space cells for the free and occupied space categorization by ray tracing. However, the space digitization create non-smooth reconstruction which limits its applications to navigation and obstacle avoidance where map resolution is less important \cite{oleynikova2016}.  
A mesh map representation provides a smooth object representation but the map update rule could be inefficient as the connections among the vertices are needed to be properly updated \cite{krainin2011}.

In \cite{bosse2009},\cite{schadler2013}, the authors presented a multi-resolution surfel map representation and fusion for a spinning 2D laser scanner. Although this representation is highly suitable for fast and robust map registration, large size surfels and its z direction volume does not make it ideal for a dense map representation.

On the other hand, a dense surfel map representation \cite{henry2012,keller2013,whelan2015,fu2018,yan2017b,behley2018rss}, originally designed to render 3D point clouds without a complicated mesh extraction step \cite{pfister2000}, is ideal for fusion, dense object representation, and rendering as shown in Figure \ref{fig:maprepresentation} (b). Thus, a surfel representation has been popular in RGB-D reconstruction problems where a rendered synthetic 2D image from the densely reconstructed object is immediately used for other purposes such as localization \cite{kerl2013} or dynamic object handling \cite{keller2013}. Despite its advantage, its introduction to LiDAR-based system has not been actively studied in the LiDAR SLAM community due to the difficulties in the data association which will be discussed in the following section.

\subsubsection{Map Element Matching}

{Data} association is one of the most important components in {map fusion.} {The simplest} way is to find a surfel with the closest distance either in Euclidean space \cite{nuchter2005} or Mahalanobis distance \cite{montemerlo2002,davison2007}. {However, generally} it is challenging to control the {map} surface resolution {without} discretizing the environment \cite{hornung2012}.

The projective data association \cite{keller2013} provides an efficient map element matching but it is restricted to a projective sensor model. Behley \etal{}~\cite{behley2018rss} applied projective data association and frame-to-model registration for the multi-beam LiDAR sensor. Their approach demonstrated that the projective data association can be successfully adapted to multi-beam LiDAR.

While approaches in this category are computationally more efficient, the frame-to-model registration method is not suitable for handling severe motion distortion and asynchronous multi-modal sensor payload (Visual-LiDAR-Inertial). Furthermore, the projective data association is limited to the LiDARs with the projective view model whereas the modern LiDAR systems \cite{tanahashi2018} often cover 360 degrees. Although, the projective data association can be imitated with LiDAR data by projecting the point cloud within the local sliding window onto a spherical image plane, this causes uneven spatial resolution where the top and bottom is dense and around the equator is sparse. Dividing the spherical image plane into an equal gird for an even sensing plane requires a nearest neighbour search, significantly reducing the efficiency of the projective data association method.

\subsubsection{Map Element Fusion}

Once matching of map elements are established, the next step is to fuse the measurements to improve estimation. {Keller \etal{} \cite{keller2013}} proposed a dense surfel fusion method by simplifying {the Bayesian estimation from 3D to 1D.} In their approach, each map element is independently updated, making its computation much simpler than {the} EKF case. {They also} utilized radial distortion as an initial uncertainty parameter and reduced the uncertainty whenever the surfel is observed again. ElasticFusion \cite{whelan2015} further extended the uncertainty as a function of sensor motion to consider the uncertainty caused by motion blur. However, those simplified Bayesian {models} are not {appropriate} for dense 3D LiDAR mapping where the existence of a surfel degeneracy often causes slower convergence.   

The approaches in this category mostly do not consider the sensor noise model \cite{whelan2015,henry2012,keller2013,behley2018rss} or they are dedicated for RGB-D models only \cite{yan2017b}.

\section{Overview}
\label{sec:Overview}

\begin{figure*}[t]
\centering{
\includegraphics[height=.175\textheight]{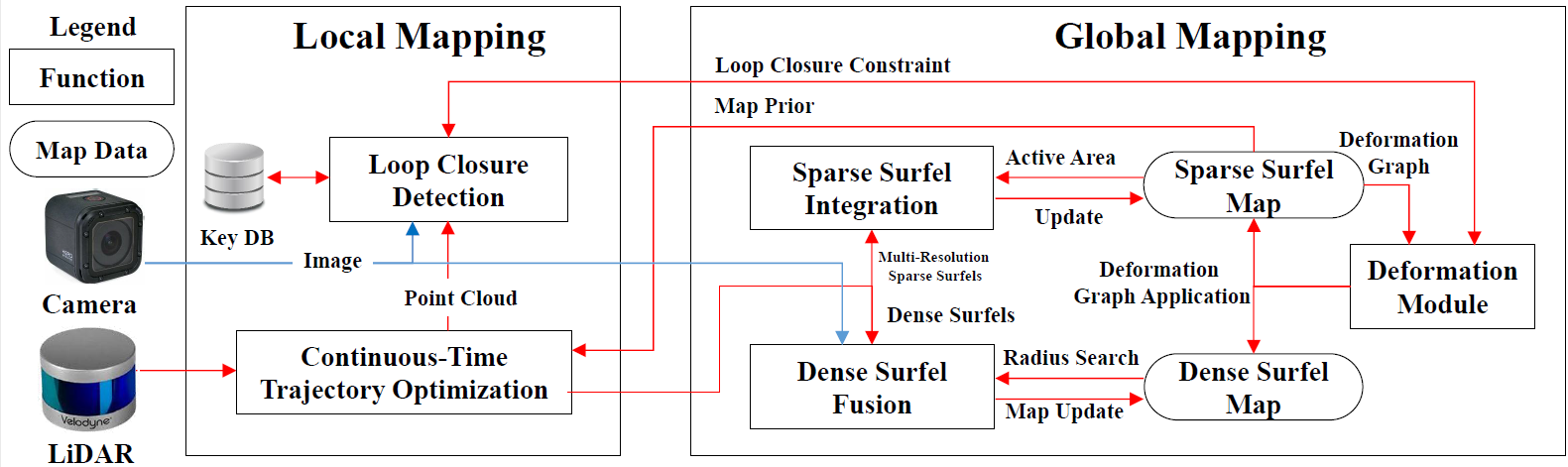}
}
\caption{
{System block diagram of our method. The device local trajectory is tracked in the Local Mapping stage, while the global consistent map is maintained in the second Global Mapping stage.}
}

\label{fig:bolckdiagram}       %
\end{figure*}

Our proposed map-centric SLAM system is composed of three main components: local mapping, global mapping and loop closure detection as shown in Figure~\ref{fig:bolckdiagram}. 
The local mapping part takes visual, IMU and LiDAR measurements to build motion-distortion-corrected map{s} using continuous-time trajectory optimization. This stage is similar to the sliding {windows in} \cite{bosse2012, bosse2009,dube2016}, but it is different in that it takes the sparse global map as a map prior. This let the new local map always registered to the global map.%
The loop closure detection module keeps generating 2D features from the visual sensor and compares them to the previously generated key frames.

The global mapping part builds and fuses surfel maps. 
Note that two different surfel maps{,} multi-resolution 3D sparse surfel map and dense surfel map, are utilized for different purposes. The multi-resolution 3D sparse surfels proposed in \cite{bosse2009} are ideal for fast and robust continuous-time trajectory optimization, whereas it is too sparse for dense representation~\cite{schadler2013}. Thus, we utilize a fixed size flat-shape hexagonal {surfels} \cite{pfister2000,whelan2016,keller2013} for dense surfel fusion.

{Global map consistency} is achieved by non-rigid deformation using the estimated misalignment from the loop closure module. It is inspired by the global deformation in ElasticFusion \cite{whelan2016}, and we further extend it by considering surfel uncertainty propagation in LiDAR. Upon a visual loop closure detection, the loop closure module finds the 6 DoF misalignment between the global map and the current frame over different places until the uncertainty of the misalignment is below a certain threshold. 
Table \ref{tbl:notations} compiles all states and parameters used in this work.

 \begin{table}[]
\centering
\begin{tabular}{ll}
\hline\noalign{\smallskip}
\textbf{Symbol}& \textbf{Description}\\
\hline\noalign{\smallskip}
&\textbf{Trajectory Optimization}\\
\hline\noalign{\smallskip}
$\textbf{T}_k$&Discrete-time trajectory representation\\

$\textbf{T}(\tau)$&Continuous-time trajectory representation\\
$\textbf{R}{(\tau)}$&Rotational component of $\textbf{T}(\tau)$\\
$\textbf{t}{(\tau})$&Translational component of $\textbf{T}(\tau)$\\

$\mathbb{Q}$ &Entire control points\\
$\textbf{Q}_k$ &A set of local control points\\ 
$\Phi(\tau)$&The indexes $k$ of local control points to recover $\textbf{T}(\tau)$\\
 
$ \textbf{e}{_{I}}$& Surfel to surfel constraint \\
$ \textbf{e}{_{M}}$&Surfel to map prior constraint\\
$ \textbf{e}{_{\alpha}}$&IMU acceleration constraint\\
$ \textbf{e}{_{\omega}}$&IMU angular velocity constraint\\
$\boldsymbol{\alpha}$ &IMU acceleration\\
$\boldsymbol{\varpi}$ &IMU rotational\\

\hline\noalign{\smallskip}
&\textbf{Map Building}\\
\hline\noalign{\smallskip}
$\mathbb{M}$&Dense surfel\\
$\mathbb{S}$&Sparse surfel\\
$\boldsymbol{\varphi}$&Surfel element\\

$\textbf{c}$&Sparse surfel centroid\\
$\boldsymbol{\Sigma}_\textbf{c}$&Sparse surfel covariance\\

$\textbf{p}$&Dense surfel centroid\\
$\hat{\textbf{n}}$&Dense surfel normal\\
$\boldsymbol{\Sigma}_\textbf{p}$&Uncertainty of a dense surfel centroid \\
$\boldsymbol{\Xi}$&Scatter matrix of a dense surfel\\

\noalign{\smallskip}\hline\noalign{\smallskip}   
\end{tabular}
\caption{Variables used to parameterize the system }
\label{tbl:notations}
\end{table}

There are two types of surfel map representations used in this paper, sparse surfel map  $\mathbb{S}_{g}$ and dense surfel map  $\mathbb{M}_{g}$. Each surfel map is individually updated with their local maps $\mathbb{S}_{l}$ and $\mathbb{M}_{l}$ which are extracted from the current laser scan.

The sparse surfel map consists of 3D ellipsoids extracted from laser points using multi-resolution voxel hashing \cite{bosse2009}. Each ellipsoid is defined with a centroid {$\textbf{c} \in \mathbb{R}^3$} and a covariance matrix {$\boldsymbol{\Sigma}_\textbf{c} \in \mathbb{R}^{3\times 3}$} which represent the distribution of points within the voxel. 
Likewise, the dense surfel map maintains the mean and covariance defining the ellipsoid of points in the surfel, which in turn allow fused estimates of position $\vec{p}\in\mathbb{R}^3$ and normal vector $\hat{\vec{n}}\in\mathbb{R}^3$.
In contrast to conventional surfels \cite{whelan2015, keller2013}, we associate uncertainty {$\boldsymbol{\Sigma}_\textbf{p}\in \mathbb{R}^{3\times 3}$} and scatter $\boldsymbol{\Xi}_\textbf{p}\in \mathbb{R}^{3\times 3}$ with the position  of each disc surfel, which are later used to merge surfels based on Bayesian filtering. The normal {$\hat{\textbf{n}}$} of surfels are extracted from the scatter matrix $\boldsymbol{\Xi}_\textbf{p}$.
Note that 3D ellipsoid surfels are expressed with ellipsoids of their covariance matrices, while 2D disc surfels are expressed with discs with normal directions.

\section{{Local Mapping}}
\label{sec:localmapping}

To deal with the asynchronous estimation and LiDAR motion distortion, we are introducing the concept of the continuous-time trajectory representation and optimization to the map-centric approach. 
In this section, we describe our proposed constraints for the map-centric local trajectory optimization that systematically couples the local continuous-time trajectory optimization method with the map-centric method. %
Also, in the later part of this section, we describe our strategy that improves the accuracy and efficiency of the trajectory optimization and update.

\subsection{Continuous-Time Trajectory Representation}
 
Let $\textbf{T}(\tau)$ be composed of translational component $\textbf{t}{(\tau}) \in {\mathbb{R}^3}$ and rotational component $\textbf{R}{(\tau)} \in {SO(3) }$  as,
\begin{equation}
\textbf{T}(\tau):=\begin{bmatrix}
 \textbf{R}{(\tau})& \textbf{t}{(\tau)}\\ 
\textbf{0}^T &1 
\end{bmatrix}.
\end{equation}

Then, utilizing the linear continuous-time trajectory representation, its value can be evaluated by an interpolation from two poses $ \textbf{T}{_k}, \textbf{T}{_{k+1}}  $ where their timestamp{s} satisfy $\tau_{k}<\tau<{\tau}_{{k+1}}$. Given poses with timestamps, their interpolation at $\tau$ is given by, 
\begin{equation}
\textbf{T}(\tau)= \textbf{T}{_{{k}}} \textbf{e}{^{\alpha[\boldsymbol{\xi}]{_\times}}},
\end{equation}
where the relative pose {$[\boldsymbol{\xi}]_\times\in\mathfrak{s}\mathfrak{e}(3)$} is defined by $\log(\textbf{T}{_{{k}}^{-1}}\textbf{T}{_{{k+1}}})$ and  the exponential mapping $\textbf{e}{^{\alpha[\boldsymbol{\xi}]{_\times}}}$ linearly interpolates the relative pose on the manifold with the interpolation ratio $\alpha = ({\tau-{\tau}_{{k}}})/({\tau_{k+1}-{\tau}_{{k}}})$.

The linear interpolation error proportionally increases when the sensor is in high-speed motion. The error is propagated to the map deteriorating the map quality. To reduce the effect of the linear interpolation error, we have generated the trajectory at 100Hz.

\begin{figure}[t]
\centering{
\includegraphics[width=.4\textwidth]{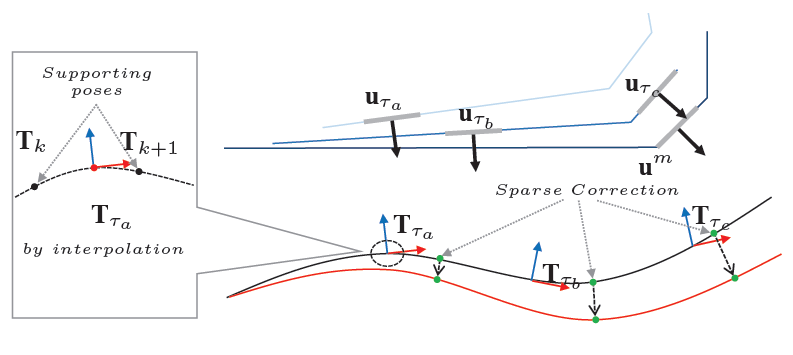}
}
\caption{
Illustration of geometrical constraints on the local trajectory. Two surfels $\textbf{u}{_{\tau_a}}$ and $\textbf{u}{_{\tau_b}}$ generated at $\tau_a$, $\tau_b$  within a local window forms constraints on the interpolated trajectory poses $\textbf{T}{_{\tau_a}}$ and $\textbf{T}{_{\tau_b}}$. The constraint between the local scan $\textbf{u}{_{\tau_c}}$ and the map prior $\textbf{u}{^m}$ forces the trajectory to be fitted onto the map prior.
}

\label{fig:ctconsts}    
\end{figure}

\subsection{Local Trajectory Constraints}

  \begin{figure}[t]
 \centering
 \includegraphics[height=.3\textheight]{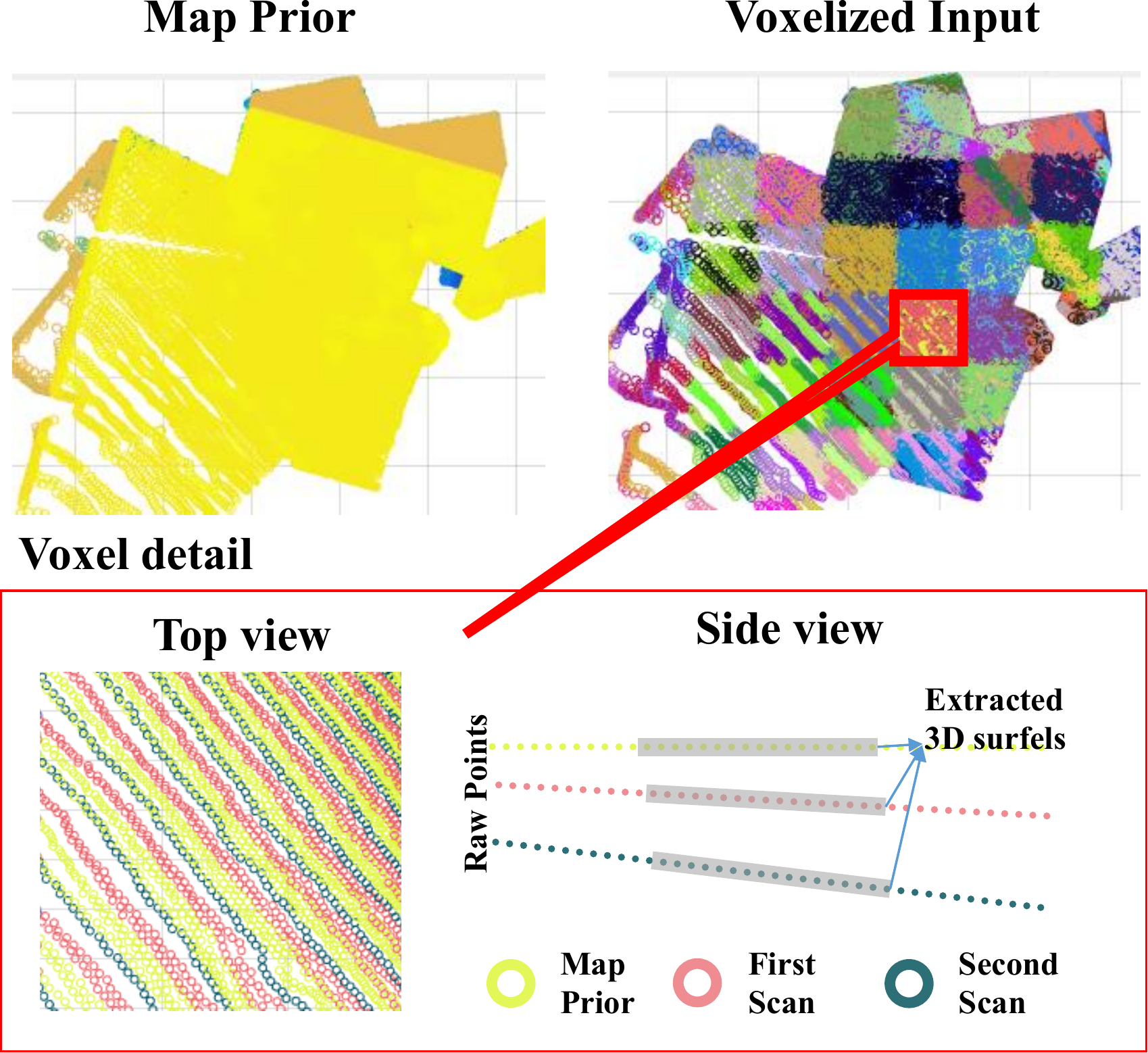}
 \caption{Visualization of the map prior and the point cloud input from LiDAR. The point cloud is voxelized and divided into different groups according to its time of generation. Extracted spatial features by multi-resolution voxels form a surfel and then utilized to find a matched surfels. Third figure shows an example of two pairs of matched surfaces constraints.}
 \label{fig:map_prior}
 \end{figure}

When {the} motion of a device is relatively moderate compared to the scanning speed and the initial trajectory guess is fairly good, shapes of local features are often well preserved in the measurements even with the motion distortion. The first two geometrical constraints utilize this property of LiDAR scans \cite{bosse2009}, \cite{zhang2015}. This process starts by transforming the LiDAR measurements with respect to the world frame and extracting {sparse ellipsoidal surfels from multi-resolution voxels} \cite{bosse2009}. Then, %
{we calculate the point-to-plane errors between two corresponding {new} surfels $a$ and $b$ in their averaged normal directions $\textbf{n}_{ab}$} as 
\begin{align}
\label{eq:surfelmat_wind}
\textbf{e}{_{I}}&=\sum_{}^{}{\Vert\textbf{n}_{ab}^T(\textbf{R}({{\tau_a}})\textbf{u}{_{\tau_a}}+\textbf{t}{({\tau_a})}-(\textbf{R}{({\tau_b})}\textbf{u}{_{\tau_b}}+\textbf{t}{({\tau_b})}))\Vert}^2 ,
\end{align}
where %
{$\textbf{u}{_\tau}$ is the centroid of a surfel, $\textbf{t}({\tau}),\textbf{R}({\tau})$ are the interpolated sensor pose at time $\tau$ from the continuous-time trajectory.}  The visualization of the constraint is given in Figure \ref{fig:ctconsts} and Figure \ref{fig:map_prior}.

{The second geometrical constraint defines the relationship between the current local map and the map prior. The map prior is obtained from the active global map which we will be described in the Session \ref{sec:fusionandmatching}. The map prior constraint makes the local trajectory always aligned to the previously built global map, which is essential for map-centric operation. Thus, we define the constraint between a {new} surfel $c$ and its corresponding {global map} surfel $m$ as }

\begin{align}
\label{eq:surfelmat_map}
\textbf{e}{_{M}}&=\sum_{}^{}{\Vert\textbf{n}_{mc}^T(\textbf{u}{^{m}}-(\textbf{R}{({\tau_c})}\textbf{u}_{\tau_c}+\textbf{t}{({\tau_c})}))\Vert}^2 ,
\end{align}
where $\textbf{n}_{mc}$ is the averaged normal directions of surfel $m$ and $c$. Note that we do not have to transform the map prior $\textbf{u}{^{ {m}}}$ because they are points in the world coordinate system. The motion distortion is corrected based on the map prior. The initial map prior is given from a short period of stationary scanning at commencement.  

On the other hand, the inertial information from IMU offers a prior on the trajectory in terms of rotational velocity and translational acceleration. The IMU measurement constraints are written as,  

\begin{align}
\textbf{e}{_{\alpha}}&=\sum_{}^{}{\Vert\boldsymbol{\alpha}_\tau-\textbf{R}{{(\tau})}^T(\frac{d^2}{d \tau^2}\textbf{t}{(\tau)}-\textbf{g})+\textbf{b}_\alpha\Vert^2} ,\\
\textbf{e}{_{\omega}}&=\sum_{}^{}{\Vert\boldsymbol{\varpi}{_\tau}-\boldsymbol{\omega}{(\tau)}+\textbf{b}_\omega\Vert}^2 ,
\label{eq:imu_traj}
\end{align}
where $\boldsymbol{\alpha}$, $\boldsymbol{\varpi}$ are respectively acceleration and rotational velocity from IMU measured at {time} $\tau$, $\textbf{g}$ is {the gravitational acceleration}, $\textbf{R}{{(\tau})},\textbf{t}{(\tau)},\boldsymbol{\omega}{(\tau)}$ are respectively interpolated rotation, translation, and rotational velocity of the continuous-time trajectory at time $\tau$, and $\textbf{b}_\alpha$, $\textbf{b}_\omega$ are the IMU biases. These two constraints form a strong restriction on local smoothness.%

\subsection{Trajectory Optimization and Update}

Compared to the number of the interpolations required for the motion distortion correction of the point cloud, the number of the interpolation for the trajectory update is much fewer.
It is not necessary to compromise accuracy for a fast interpolation by using linear interpolation for this stage.
Therefore, we utilize B-Spline for the smooth update.

We start by defining a set of control points $\textbf{Q}_k \in \mathbb{Q}$ which is created at an interval of $\tau{_{step}}$ and longer than the discrete trajectory sampling interval. The trajectory corrections are achieved by finding a set of parameters that {minimize the residuals of Eq. (\ref{eq:surfelmat_wind})-(\ref{eq:imu_traj})} as,
\begin{equation}
\hat{x}=\underset{x}{\mathrm{argmin }}{\ \textbf{e}{_{I}}+\textbf{e}{_{M}}+\textbf{e}{_{\alpha}}+\textbf{e}{_{\omega}}},
\end{equation}
where $x=[\mathbb{Q},\textbf{b}_\omega,\textbf{b}_\alpha,d]$. $d$ is a time lag parameter and $\mathbb{Q}$ is control points.

Then, the correction to the discrete trajectory is made by,
\begin{equation}
\textbf{T}'_k= d\textbf{T}(\tau_k,\textbf{Q}_{k\in \Phi(\tau_k)})\textbf{T}{_{{k}}},
\end{equation}
where the correction $d\textbf{T}$ for each discrete trajectory at $\tau_k$ is defined as an interpolation of neighboring trajectory elements {$\textbf{Q}_{k \in \Phi(\tau)}= \left\lbrace \mathbf{c_{t}}_{k-1}, \mathbf{c_{r}}_{k-1},\cdots \mathbf{c_{t}}_{k+2}, \mathbf{c_{r}}_{k+2} \right\rbrace $}. $\Phi(\tau)$ represents a set of neighbor indices at $\tau$. Although the utilized  control points are in Euclidean space as $\mathbf{c_{r}}_k,\mathbf{c_{t}}_k \in {\mathbb{R}^3} $, in each iteration they always starts from zero. Thus, it is free from the singularity problem.

For the recovery of the correction $d\textbf{T}$ from the control points we have utilized the B-spline as,
\begin{equation}
\textbf{t}{_\tau}= \begin{bmatrix}
t^3 &t^2  & t & 1 
\end{bmatrix}
\frac{1}{6} 
\begin{bmatrix}
1 & -3 & 3 &-1 \\
0 & 3 & -6 &3 \\ 
0 & 3 & 0 &-3 \\ 
0 & 1 & 4 &1 
\end{bmatrix}
\begin{bmatrix}
\mathbf{c_{\textbf{t}}}_{k-1}\\ 
\mathbf{c_{\textbf{t}}}_{k}\\ 
\mathbf{c_{\textbf{t}}}_{k+1}\\ 
\mathbf{c_{\textbf{t}}}_{k+2}
\end{bmatrix},
\end{equation}
where $t \in [0,1]$. In a similar way, rotational component $\textbf{r}{_\tau}\in\mathfrak{s}\mathfrak{o}(3)$ is defined from separate control points $\mathbf{c_{r}}$ and converted to ${SO(3) }$ by an exponential mapping $\textbf{R}{_\tau}=\textbf{e}{^{[\textbf{r}{_\tau}]{_\times}}}$.

The approximation method in \cite{bosse2012} utilized $SO(3) +\mathbb{R}^3$ update scheme where the translation update is given as ($\textbf{t}'= \delta\textbf{t}+\textbf{t}$). 
The selection of $SO(3) +\mathbb{R}^3$ over $SE(3)$ was reasonable as the $SE(3)$ update with ($\textbf{t}'= \delta\textbf{t}+\textbf{e}^{[\boldsymbol{\omega}]_\times}\textbf{t}$) could cause an accuracy problem with linearization. Especially when $\textbf{t}$ is large, the linearization error of the rotation enlarges the translation error in the second term $\textbf{e}^{[\boldsymbol{\omega}]_\times}\textbf{t}$.
Separately updating $SO(3) +\mathbb{R}^3$ reduces the linearization error in the large map global optimization. However, $SO(3) +\mathbb{R}^3$ results in suboptimal. %
Considering that our approach does not need a global trajectory optimization, we use a $SE(3)$ update. Also, as the map grows, this problem can be solved by simply converting the trajectory and estimations in the local window to be relative to the first frame in the local window.

After the optimization, local dense and sparse surfel maps $\mathbb{M}_{l}$, $\mathbb{S}_{l}$ are built and updated by the optimized trajectory. 
Each dense surfel in the local dense map is composed of position $\textbf{p}\in \mathbb{R}^3$, surfel normal $\textbf{n}\in \mathbb{R}^3$, and timestamp $t$. Also, the uncertainties of position and scatter matrix ${\Sigma }{_\textbf{p}} \in \mathbb{R}^{3\times 3}$, $\boldsymbol{\Xi} \in \mathbb{R}^{3\times 3}$, which are utilized in dense surfel matching and fusion are calculated from their neighboring points.
On the other hand, local sparse surfel maps $\mathbb{S}_{l}$ which are utilized in Equation (\ref{eq:surfelmat_wind}), (\ref{eq:surfelmat_map}) are updated by the optimized trajectory. Each sparse surfel is defined with a centroid $\textbf{c} \in \mathbb{R}^3$ and a covariance matrix $\boldsymbol{\Sigma}_\textbf{c} \in \mathbb{R}^{3\times 3}$ which represents the distribution of points within the voxel. The local maps $\mathbb{M}_{l}$, $\mathbb{S}_{l}$ are fused into the global maps $\mathbb{M}_{g}$, $\mathbb{S}_{g}$ in the following section.

\section{Dense Surfel Matching and Fusion}
 \label{sec:fusionandmatching}

In this section, we introduce our probabilistic surfel fusion. We will detail our two staged probabilistic surfel matching that can maintain surface resolution of the dense map to a specified value. Also, a new probabilistic surfel representation that can correctly track the surfel normal under the existence of degeneracy will be highlighted.

\subsection{Surfel Uncertainty Modeling}
\label{sec:surfel_uncertainty}
\subsubsection{{Uncertainty in Position and Orientation}}
We model the surfel position and shape as a random ellipse using a normal inverse Wishart model, based on the statistical model of LiDAR sensor noise measurements~\cite{pomerleau2012}. 
If the LiDAR points were noise-free, the normal inverse Wishart model would provide an exact closed-form estimate of the mean and covariance matrix from which the points were drawn. In practice, the positional uncertainty of the LiDAR points is higher along the beam direction. Therefore, we adopt the method of \cite{FelFra11}, which provides estimates under non-homogeneous noise matrices. By estimating both the mean and covariance matrix, we recursively integrate knowledge on both the position and orientation of the surfel.

Surfels are modelled as a Gaussian distribution of points with mean and covariance to be estimated. The LiDAR uncertainty is an additional Gaussian measurement noise, such that the distribution of each point is  $\mathcal{N}\{\mathbf{z};\boldsymbol{\mu},\boldsymbol{\Sigma}+\textbf{Q}_w\}$, where $\boldsymbol{\mu}$ and $\boldsymbol{\Sigma}$ are the unknown surfel mean and covariance, and $\textbf{Q}_w$ is the measurement noise in world coordinates. Since points in the same surfel that are local in time will have the same geometry, this can be modelled as
\begin{equation}\label{eq:SurfelNoiseCovariance}
\textbf{Q}_w ={^w}\textbf{R}_{l}{^l}\textbf{R}_{b}{\textbf{Q}_b}({^w}\textbf{R}_{l}{^l}\textbf{R}_{b})^\top ,
\end{equation}
where ${}^w\mathbf{R}_l$ and ${}^l\mathbf{R}_b$ are rotation matrices from laser to world coordinates, and from beam to laser coordinates, respectively. The amount of uncertainty along the beam direction is defined by the sum of {the} distance uncertainty {$\sigma_r^2$} and additional uncertainty {$\sigma_i^2$} caused by {the} incident angle {\cite{kneip2009,okubo2009}}. The uncertainty covariance in beam coordinates is $\textbf{Q}_b = {\rm diag}(\sigma_r^2, \sigma_r^2, \sigma_i^2 + \sigma_d^2)$, where $\sigma_r^2$ is noise due to the beam radius, $\sigma_d^2$ is the nominal depth variance following the model of \cite{pomerleau2012}, and $\sigma_i^2$ is an additional uncertainty caused by the incidence angle \cite{kneip2009,okubo2009}. Surfel observations from different times are assumed conditionally independent given the surfel mean and covariance.

\begin{figure*}[t]
\centering{
\subfloat[]{\def\svgwidth{78mm} 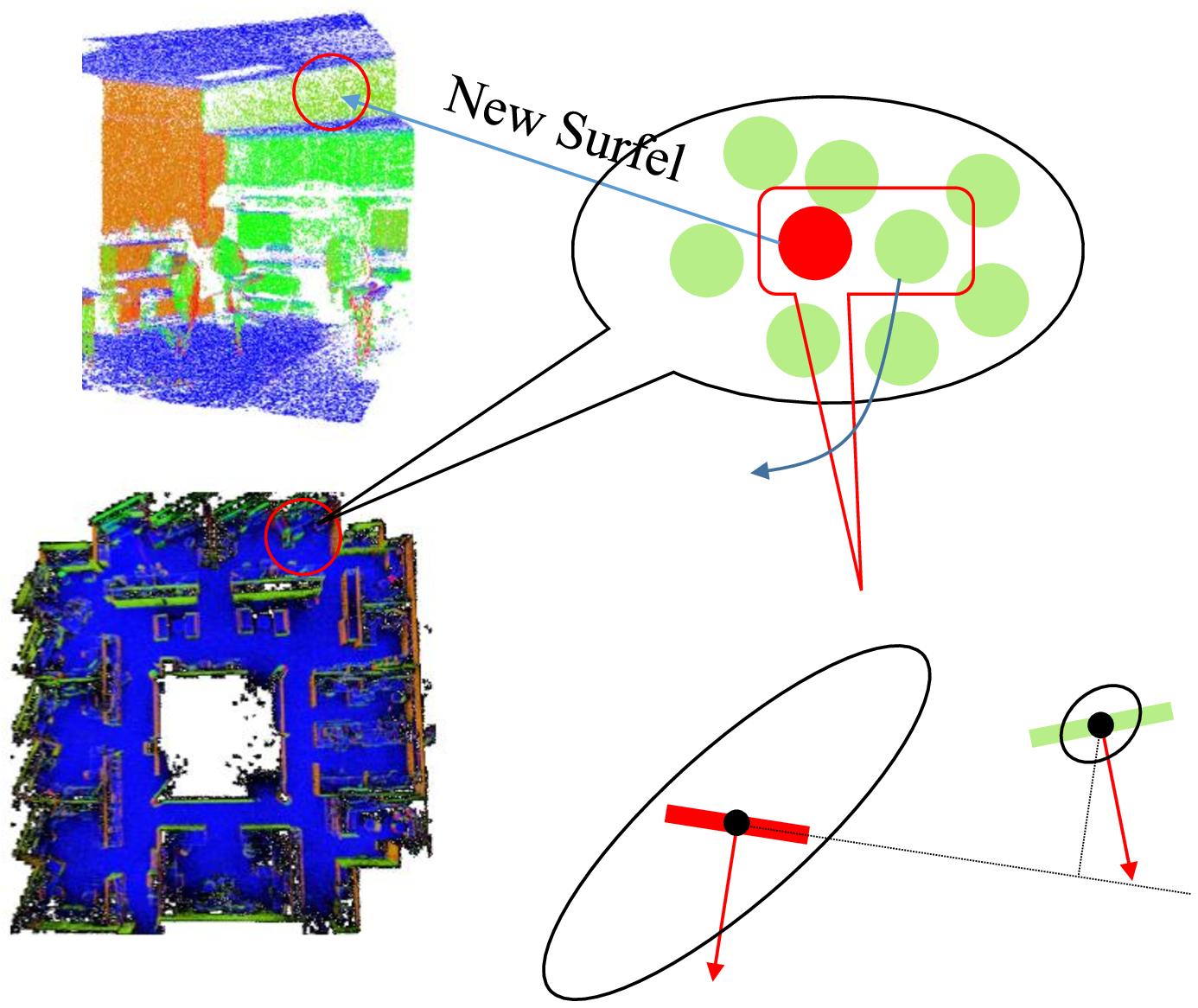}~
\subfloat[]{\includegraphics[height=.25\textheight]{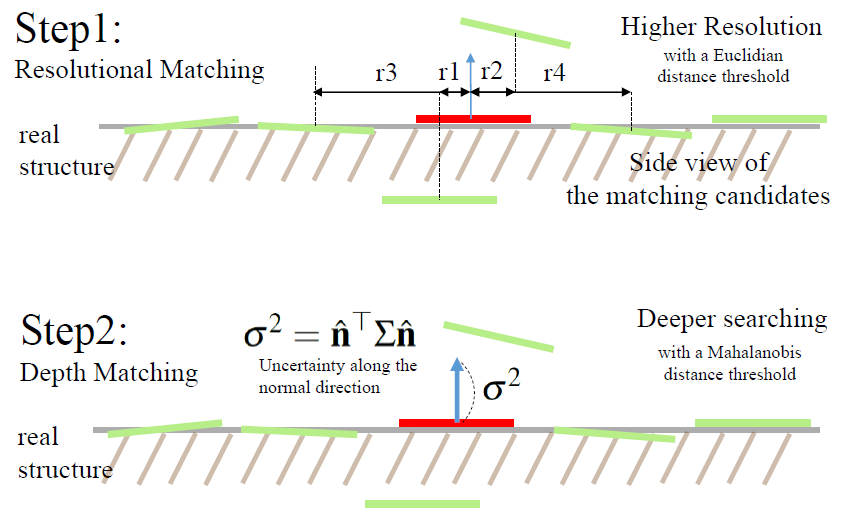}}
}
\caption{
(a) Illustration of surfel matching between a local map surfel and the global map surfels. (b) Proposed two staged matching algorithm. The first step controls map resolution whereas the second step reduces map noise by searching deeper along the LiDAR beam direction.
}
\label{fig:2stage_matching}       %
\end{figure*}

\subsubsection{{Uncertainty in Colour}}

The local point cloud is rendered to the image frame for visualization. We parameterize three uncertainty factors that affect the accuracy of the rendered point cloud  as
$\alpha {_{r}}= e(r/r_{th})-0.5, \alpha {_{v}}= f \sigma([d_c, d_{d},d_{u},d_{l},d_{r}])-0.5, \alpha {_{d}}= gd_c-0.5$, where $\alpha {_{r}}$ penalizes the projected points around edge of the image by the distance from the optical center $r$ considering the wide angle lenses often show low quality image around the edge. $\alpha {_{v}}$ checks if the projected feature is around the object edge by looking at the neighboring depth $d_{d},d_{u},d_{l},d_{r}$. Finally, $\alpha {_{d}}$ gives more weight to the close object to the camera by checking the depth $d_c$. Combining the attributes in a sigmoid function, the colour uncertainty $\sigma {_{c}}$ is given by,   
\begin{align}
\sigma {_{c}}&=(1+e^{-w(\alpha {_{r}}+\alpha {_{v}}+\alpha {_{d}})})^{-1} .\end{align}

\subsection{Surfel Matching}
\label{sec:surfel_matching}

The {local} surfels {$\mathbb{M}_{l}$ are} transformed into world {coordinates} to find matched surfels in the global dense surfel map $\mathbb{M}_{g}$. The matching {process begins} with finding a set of candidate surfels {$\mathbb{A}_g$} {for each surfel $\boldsymbol{\varphi}_l \in \mathbb{M}_{l}$}. For efficient matching, initial matching candidates are selected by the octree-based nearest neighbor search algorithm.

{Then}, the resolutional distance $r$ {between} each source and destination pair in Figure \ref{fig:2stage_matching} {is} compared with a resolution threshold {$\theta_r$} to decide if {their projections} are close enough {on} the surface. 
{If so,} we check the depth $d$ in the Mahalanobis distance. To consider the uncertainty only along the normal direction, we propagate the positional 3D uncertainties of source and destination surfel $\Sigma_d$, $\Sigma_s$ into 1D along each normal direction by {$\sigma^2 = \hat{\textbf{n}}^\top\boldsymbol{\Sigma}\hat{\textbf{n}}$}. 

Finally, {if the 1D Mahalanobis distance along the surface normal direction is less than a threshold $\theta_d$, we assume that they are in correspondence, and put the matched surfel into $\mathbb{B}_g$.} {Note that the resolutional distance $d$ is compared in the Euclidean space to preserve the desired surface resolution in the Euclidean space. }%
{Algorithm \ref{alg:Match} summarizes this surfel matching process.}
Note that our matching method enables the matching process to search more along the beam direction, while effectively maintaining the desired surface resolution without a voxel grid. 

The RGB values of each matched point in the image are found by rendering the surfel map at each camera position.

\begin{algorithm}[!t]
\SetAlgoLined
\setcounter{ALG@line}{1}

\KwIn{{Global dense surfel map $\mathbb{M}_{g}$ and a surfel $\boldsymbol{\varphi}_l \in \mathbb{M}_{l}$} }
\KwOut{A set of matched surfels {$\mathbb{B}_g \subseteq \mathbb{M}_{g}$}}

        {$\mathbb{A}_g \leftarrow OctreeSearch(\boldsymbol{\varphi}_l,\mathbb{M}_{g})$}%
        
        \ForEach{{$\boldsymbol{\varphi}_g \in \mathbb{A}_g$}}{%
                {$[\ r, d\ ] \leftarrow Point2PlaneDist(\boldsymbol{\varphi}_g, \boldsymbol{\varphi}_l)$}
                
                \If{{$r < \theta_r$}}{                        
                        {$\sigma^2 = \hat{\textbf{n}}_{s}^\top\boldsymbol{\Sigma}_s\hat{\textbf{n}}_{s}+\hat{\textbf{n}}_{d}^\top\boldsymbol{\Sigma}_d\hat{\textbf{n}}_{d}$}
                        
                        \If{ {$d/\sigma < \theta_d$}}{
                                {$\mathbb{B}_g \leftarrow \mathbb{B}_g \cup \boldsymbol{\varphi}_g$}
                        }
                }
        }
\caption{{Finding Surfel Matches}}
\label{alg:Match}
\end{algorithm}

\subsection{Surfel {Fusion}}

As described previously, surfels are modelled as a Gaussian distribution of points with mean $\boldsymbol{\mu}$ and covariance $\boldsymbol{\Sigma}$. These parameters are estimated recursively using a normal inverse Wishart model. In each update, the LiDAR points $\{\mathbf{z}_i\}$ are summarised via the noise uncertainty $\textbf{Q}_w$ given in \eqref{eq:SurfelNoiseCovariance}, alongside the number of points $n$, the measurement mean and the scatter matrix:
\begin{align}
\bar{\mathbf{z}} &= \frac{1}{n}\sum_{i=1}^n \mathbf{z}_i \\
\bar{\mathbf{Z}} &= \sum_{i=1}^n (\mathbf{z}_i-\bar{\mathbf{z}})(\mathbf{z}_i-\bar{\mathbf{z}})^T
\end{align}
As described further in Appendix A, the method in Algorithm 2 recursively updates the surfel parameters, based on the random matrix approach in \cite{FelFra11}. By directly estimating the surfel ellipsoids, surface centroids and normals are estimated through an integrated Bayesian model. Normal estimates may then be extracted from the estimated ellipsoid covariance via eigen decomposition. This is different from our previous work \cite{park2017b,park2017c} and other approaches \cite{yan2017b}, where normal vectors were extracted from each surfel, and fused by assuming an error model on these estimates. Our new approach
uses an error model that is implied by the single point error model.

The colours are fused similarly with a Gaussian distribution but in $\mathbb{R}$ as there is no measurement to estimate the uncertainty of each colour channel. 
The surfels that are not matched to the global map will be added to the global map as a new unstable surfel. To effectively remove the surfels generated from LiDAR data with non-Gaussian noise, any unstable surfels that are not re-observed on revisits will be deleted from the global map.

\begin{algorithm}[!t]
\SetAlgoLined
\KwIn{Map surfel $\boldsymbol{\varphi}_{i}^{d}$, Input surfel $\boldsymbol{\varphi}_{i}^{s}$}
\KwOut{Updated map surfel $\boldsymbol{\varphi}_i$}
        \ForEach{pair of matched surfels $\boldsymbol{\varphi}_{i}^{d}$, $\boldsymbol{\varphi}_{i}^{s}$}{%
                Centroid Fusion:\

                $\boldsymbol{\Sigma} {_d} {\leftarrow} \boldsymbol{\Sigma} {_d}-\textbf{K}\boldsymbol{\Sigma} {_d}$

                $\textbf{p}_{d} {\leftarrow} \textbf{p}_{d}+\textbf{K}(\textbf{p}_{s}-\textbf{p}_{d}) $

                $\boldsymbol{\Xi}_{d} {\leftarrow} \boldsymbol{\Xi}_{d}+\bar{\textbf{N}} +\bar{\textbf{Y}} $
                
                Normal Extraction:\

                $[\textbf{v},\ \textbf{D}] \leftarrow eigen\_decomposition\textbf(\boldsymbol{\Xi}_{d} )$

                $\hat{\textbf{n}}_d = \textbf{v}_{3}$

                Colour Fusion:\
                
                $\sigma {_d} {\leftarrow} (\sigma {_{s}^{-1}}+\sigma {_{d}^{-1}})^{-1}$

                $\textbf{g}_{d} {\leftarrow} \frac{ (\sigma {_d^{-1}}\textbf{g}{_d}+\sigma {_s^{-1}}\textbf{g}{_s})}{(\sigma {_{s}^{-1}}+\sigma {_d^{-1}})}$
               
                }

\caption{Surfel Fusion}
\label{alg:fusion}
\end{algorithm}

\subsection{Active and Inactive {Maps}} 
One of the most important assumptions in the fusion is that the global map consistency should be guaranteed around the area where a fusion occurs. %
In {the} worst case, surfels will be fused right before the loop closure. The fusion of the surfels with a misalignment will make an unwanted deformation in the local area. Thus, similar to \cite{whelan2016}, we introduce active and inactive maps $\mathbb{A}$, $\mathbb{I}$  according to the surfel timestamps so that new surfels are always optimized and fused within the active area. To detect the misalignment between the active and inactive areas, it finds the amount of overlap and misalignment by rigid Iterative Closest Point (ICP). For a robust and fast ICP, only sparse surfels are utilized with a geometrical weight. When the overlap and misalignment is large enough after the ICP, it triggers deformation that will be described in the next section. To maintain the map coherency between active and inactive components, matched inactive components to active components are found and updated in every step. Algorithm \ref{alg:tempfusion} shows our temporal fusion method for LiDAR surfels.

\begin{algorithm}[!t]
\SetAlgoLined

\KwIn{{New Frame $\mathbb{S}_{l}$, $\mathbb{M}_{l}$, Global Maps $\mathbb{S}_{g}$, $\mathbb{M}_{g}$} }
\KwOut{{Updated Global Maps $\mathbb{S}_{g}$, $\mathbb{M}_{g}$} }

        $\mathbb{A} \leftarrow GetCurrentActive(\mathbb{S}_{g},\mathbb{M}_{g})$
        
        $\mathbb{I} \leftarrow GetCurrentInactive(\mathbb{S}_{g},\mathbb{M}_{g})$

        {$\mathbb{S}_{g},\mathbb{M}_{g} \leftarrow SurfelFusion(\mathbb{S}_{l},\mathbb{M}_{l},\mathbb{A})$}
        
        {$[\ \textbf{R}, \textbf{t}, inlier, dist\ ] \leftarrow ICP(\mathbb{S}_{l},\mathbb{I})$}

        \uIf{$inlier > \theta_{in} \ \& \ dist > \theta_{d}$}{
        
                $[\hat{\mathbf{R}}_j,\hat{\mathbf{t}}_j] \leftarrow GraphOptimization(\textbf{R}, \textbf{ t},\mathbb{S}_{l},\mathbb{S}_{g})$
                
                $\mathbb{S}_{g},\mathbb{M}_{g} \leftarrow MapDeform(\hat{\mathbf{R}}_j,\hat{\mathbf{t}}_j,\mathbb{S}_{g},\mathbb{M}_{g})$
        }
        \Else{
        
                {$n \leftarrow NearistNeighborSearch(\mathbb{A},\mathbb{I})$}
                
                \If{$n < \theta_{n}$}{
        
                        {$\mathbb{S}_{g} \leftarrow Fusion(\mathbb{A},\mathbb{I})$}
                }
        }

\caption{{Temporal Surfel Fusion}}
\label{alg:tempfusion}
\end{algorithm}

\section{Global Map Building}
\label{sec:deformation} 
As the proposed system does not maintain any trajectory, an alternative method for a loop closure is necessary.   
In this section, we describe our approach
for maintaining the map to be globally consistent where we elastically deform the map upon the loop closure detection. Also, we highlight our sequential metric localization method that can stably and robustly estimate the misalignment.

\subsection{Map Deformation}

Upon loop closure detection, map deformation is carried out to maintain global map consistency. Deformation nodes and loop closure constraints are only selected from the sparse surfel map $\mathbb{S}_{g}$. Then, once the optimal deformation is found by an optimization, it is applied to both entire map elements in $\mathbb{M}_{g}$, $\mathbb{S}_{g}$. We adopted the graph deformation technique for the introduction of elasticity in the map and temporally connected deformation graph of \cite{whelan2016}. {However,} our approach is different from \cite{whelan2016} in that the number of deformation nodes is decided by the area of the reconstructed space and the uncertainty is also deformed along with the normal and centroid.

\subsubsection{Graph Construction} 

Graph nodes are constructed from randomly selected surfel centroids of $\mathbb{S}_{g}$. The centroids are selected to uniformly represent the space. 
The number of nodes is decided by the area of the space. As the proposed dense surfel map fusion maintains a canonical form of surface without redundancy, the area of the map surface, $A$ can be easily found by counting the number of total surfels. Thus, considering that each surfel of the map has a fixed radius circle, the total number of nodes $n$ is,

\begin{equation}
n = \left \lceil{\gamma\pi r^2(E/A) }\right \rceil ,
\end{equation}
where $\gamma$ is the total number of map elements, $r$ is the radius of the circle, and $E$ is the number of nodes in the area. Denser nodes help reduce local spatial distortion but exponentially increase the computational cost. To reject any temporally uncorrelated connection, we order the nodes by temporal sequence and define their neighbors according to their temporal closeness. The surfel timestamps are updated to the current time when they are revisited.

The deformation graph is composed of node positions $\textbf{g}{_j} \in \mathbb{R}^3$, node rotations and translations, $ \textbf{R}{_j},\textbf{t}{_j} \in {SE(3)}$, and a set of neighbors $\mathbb{V}(\textbf{g}{_j})$. Most of the time, the map is deformed by the difference between node translations. 
Regarding the number of neighbors, previous research \cite{whelan2016} {empirically} {indicates} that four neighbor nodes are sufficient.

\subsubsection{Graph Deformation} 
Assuming that a set of graph node locations $\textbf{g}{_j}$ are established and their deformation attributes $ \textbf{R}{_j},\textbf{t}{_j}$ are found, the graph deformation can be applied to the entire map. Given a set of node parameters, an influence function $\phi (\textbf{p}{_i})$ that deforms any given point $\textbf{p}{_i}$ by the $j$-th node $\textbf{g}{_j}$ is defined as,
\begin{equation}
\phi (\textbf{p}{_i})=\textbf{R}{_j}(\textbf{p}{_i}-\textbf{g}{_j})+\textbf{g}{_j}+\textbf{t}{_j}.
\end{equation}

For a smooth blending of deformation, the complete deformation of a point is defined as a weighted sum of the influence function $\phi (\textbf{p}{_i})$ with its neighbor nodes $\mathbb{N}(\textbf{p}{_i})$ around the position. When selecting nodes, their temporal and spatial closeness are considered. Thus, the deformed pose $\textbf{p}{_i^{'}}$ is defined as, 
\begin{equation}
\textbf{p}{_i^{'}}=\sum_{j\in \mathbb{N}(\textbf{p}{_i})}^{}{w{_j}(\textbf{p}{_i})\phi (\textbf{p}{_i})},
\label{eq:deformation}
\end{equation} 
where the weight $w{_j}(\textbf{p}{_i})$ is decided by the distance between the node $\textbf{g}{_j}$ and the point $\textbf{p}{_i}$,
\begin{equation}
w{_j}(\textbf{p}{_i})=(1-\left \|\textbf{p}{_i}-\textbf{g}{_j}  \right \|/d{_{max}}),
\end{equation}
where $d{_{max}}$ represents the max distance between the node and point within $\mathbb{N}(\textbf{p}{_i})$. Those nodes that are relatively far from the given point $\textbf{p}{_i}$ have smaller weights and less effect. Note that Equation (\ref{eq:deformation}) will be applied to both sparse and dense maps $\mathbb{S}_{g}$, $\mathbb{M}_{g}$.

Not only surfel centroids, but other surfel attributes also need to be deformed. The new normal direction of the surfel is defined as,
\looseness=-1
\begin{equation}
\textbf{n}{_i^{'}}=\sum_{j \in \mathbb{N}(\textbf{p}{_i})}^{k}{w{_j}(\textbf{p}{_i})\textbf{R}{_j^{-T}}\textbf{n}{_i}}.
\label{eq:normal}
\end{equation}

The uncertainty and geometry matrices ${\Sigma }{_\textbf{p}{_i}}$, $\boldsymbol{\Xi}{_i}$, ${\Sigma }{_\textbf{c}{_i}}$  are deformed by first order linear propagation. 
The general form of the propagated uncertainty and geometry in the new deformed space is given as,
\begin{equation}
{\Sigma' }{_i} = \textbf{R}{^{'}_i}{\Sigma }{{_i}}\textbf{R}{^{'T}_i},
\label{eq:covmatrix}
\end{equation}
where $\textbf{R}{^{'}_i}$ represents the blended rotation of the surfel location and defined as,
\begin{equation}
\textbf{R}'{_i}=\sum_{j\in \mathbb{N}(\textbf{p}{_i})}^{k}{w{_j}(\textbf{p}{_i})\textbf{R}{_j}}.
\end{equation}

\subsubsection{Graph Optimization} 
In this section, we will describe constraints for the graph optimization, given a set of matched surfel centroid pairs $\textbf{p}{_{src}}$, $\textbf{p}{_{dest}} \in \mathbb{S}_{g}$. During the optimization, the transformation $ \textbf{R}{_j},\textbf{t}{_j}$ of each node which minimize the source and destination will be found. The first constraint that reduces the difference between the deformed source sets $\textbf{p}'{_{src}}=\phi (\textbf{p}{_{src}})$ and target surfel $\textbf{p}{_{dest}}$ is,

\begin{equation}
\textbf{e}{_{loop}}=\sum {\left \|  \textbf{p}'{_{src}}-\textbf{p}{_{dest}} \right \|{^2} }.
\end{equation}

The distortion is applied over all the global map surfels including the destination surfels $\textbf{p}{_{dest}}$ itself. To prevent the infinite loop of deformation between source and destination, the following pinning constraint will fix the destination surfels not to be deformed,

\begin{equation}
\textbf{e}{_{pin}}=\sum {\left \|  \textbf{p}'{_{dest}}-\textbf{p}{_{dest}} \right \|{^2} }.
\end{equation}

To guarantee the smooth deformation over the whole region, the following term spreads the deformation to the neighboring nodes. When the node rotation $\textbf{R}{_j}$ is near the identity matrix, the smoothness term indicates the distance between $\textbf{t}{_j}$ and its neighboring node translation $\textbf{t}{_k}$,
\begin{equation}
\textbf{e}{_{reg}}=\sum_{j}^{m}{\sum_{k \in \mathbb{V}(\textbf{g}{_j})}{\left \| \textbf{R}{_j}(\textbf{g}{_k}-\textbf{g}{_j})+\textbf{g}{_j}+\textbf{t}{_j}-\textbf{g}{_k}-\textbf{t}{_k} \right \|}{^2}}.
\end{equation}

Then, the optimal node transformation $\hat{\mathbf{R}}{_j},\hat{\mathbf{t}}{_j}$ that minimizes radical deformation and loop closing error is defined as,
\begin{equation}
[\hat{\mathbf{R}}{_j},\hat{\mathbf{t}}{_j}]=\underset{\textbf{R}{_j},\textbf{t}{_j} \in  SE(3)}{\mathrm{argmin}}{\omega {_{reg}}\textbf{e}{_{reg}}+\omega {_{pin}}\textbf{e}{_{pin}}+\omega {_{loop}}\textbf{e}{_{loop}}},
\end{equation}
where $\omega {_{reg}}$, $\omega {_{pin}}$, $\omega {_{loop}}$ represent the weights for each constraint to tune elastic property. {We follow the proposed values in \cite{whelan2016}}. %

Since the cost function is a non-linear pose graph optimization problem in the form of $f(\textbf{T})=\textbf{Tp}+\textbf{c}$ on a manifold, we use {the} non-linear iterative Gauss-Newton method.

\subsubsection{Global Loop Closure} 
There are two sources of loop closure detection. For a loop where the source and destination distance is moderately close, the ICP in Algorithm \ref{alg:tempfusion} will detect the misalignment first. However, for large-scale mapping where large misalignment can occur, an alternative place recognition such as \cite{bosse2013}, \cite{dube2017} is required. We propose a robust method for place recognition and metric localization using both visual and geometrical information, which will be detailed in the following section. In both cases, loop closure constraints are given as follows,   
\begin{equation}
\textbf{p}{_{dest}} = \textbf{R}\textbf{p}{_{src}}+\textbf{t},
\label{eq:loop_closure}
\end{equation}
where $\textbf{p}{_{src}}$ are randomly selected points from $\mathbb{S}_{l}$ and $\textbf{R},\textbf{t}$ denote misalignment between $\mathbb{S}_{l}$ and $\mathbb{I}$. 
Note that we utilize a transformed destination $\textbf{p}{_{dest}}$ from the $\textbf{p}{_{src}}$ to prevent any unwanted deformation caused by source and destination point difference. This way, the deformation {implicitly reduces the misalignment $[\textbf{R},\textbf{t}]$ to be $[\textbf{I}, \textbf{0}]$}.

\begin{table*}[]
\centering
\begin{tabular}{lccccc}
\hline\noalign{\smallskip}
Model&  \multicolumn{2}{c}{Composition Model}  &  \multicolumn{3}{c}{Approximation Model} \\
 \cline{2-6}
Interpolation& Linear & \multicolumn{1}{c}{Linear $\mathfrak{s}\mathfrak{e}(3)$} & Spline  & Spline& Spline\\
 \cline{2-6}
Optimization Model&  \multicolumn{1}{c}{Linear Composition} & \multicolumn{1}{c}{Spline Composition} & Spline Direct  & Spline Direct& Spline Direct\\
 \hline\noalign{\smallskip}
Update Method   &        $SO(3) +\mathbb{R}^3$ & $SE(3)$    & $SE(3)$   & $SE(3)$ & $SE(3)$  \\ 
No. of Compositions/Controls    &11 &  11 &11&  51&  101   \\
No. of Trajectory Poses    &500  & 500 &N/A&  N/A&  N/A  \\
No. of States    &66 & 66  &66&  306&  606  \\
Final $\textbf{t}$ Accuracy ($mm$)        &39.0&\textbf{10.3} &103.0&21.8&23.1\\
Final $\textbf{r}$ Accuracy ($10^{-3}$ rad)        &5.0&\textbf{1.2} &93.7&10.8&5.1\\
Reference        & \cite{bosse2012}  &  Ours        & \cite{furgale2012}&\cite{furgale2012}&\cite{furgale2012}\\
\noalign{\smallskip}\hline\noalign{\smallskip}   
\end{tabular}
\caption{Comparison of different continuous-time trajectory optimization in a simulation. Our method has significantly lower state dimension, compared to the same motion resolution in \cite{furgale2012} but also higher accuracy than \cite{bosse2012}.}
\label{tbl:simulated_results_CTSLAM}
\end{table*}

\subsubsection{Misalignment Estimation for Global Loop Closure} 

Disadvantage of the map-centric method is that the result of an incorrect loop closure is more destructive and is not reversible. Thus, for reliable estimation of the misalignment $[\textbf{R},\textbf{t}]$ even with incorrect initial guesses and large drifts, we tightly fuse the surfel-based point-to-plane constraint with the 3D feature based point-to-point constraint as,   

\begin{equation}
\label{eq:rigidICP}
\textbf{e}{_{p}}=\textbf{n}^\top(\textbf{p}_r-(\textbf{R}\textbf{p}_s+\textbf{t})),
\end{equation}
\begin{equation}
\label{eq:ICP}
\textbf{e}{_{f}}=\textbf{p}'_r-(\textbf{R}\textbf{p}'_s+\textbf{t}),
\end{equation}
where the pair of the matched surfel centroids $\textbf{p}_r$, $\textbf{p}_s$ {are} acquired by the nearest neighbour method in {the} surface normal and centroid space. The matched sparse 3D features $\textbf{p}'_r$, $\textbf{p}'_s$ are found through 3D feature matching such as FPFH \cite{rusu2009}, but not limited by the types of feature descriptor. The surfel based constraint and the 3D feature based constraint are complementary to each other. While the surfel constraint lacks semantic correspondence which result in sub-optimal solution, the 3D feature often suffers from noisy LiDAR point cloud or partial observation. When combined, the 3D features roughly guide the optimization to where the surfel correspondence can be correctly found and surfel constraints refine the final result. 
Similar to our previous approach \cite{park2019,guo2019a}, we sequentially estimate $[\textbf{R},\textbf{t}]$ at different places over time which greatly reduces the chance of failure.
Refers to the Appendix B for the details of the sequential pose fusion.

\section{Experiments}
\label{sec:experiments}

In this section, we present the quantitative and qualitative performance analysis of the proposed method in terms of trajectory estimation and reconstruction quality with various sensor payload configurations and environments. Also, we provide a qualitative comparison of the metric localization accuracy with various approaches. 

We have utilized CT-SLAM \cite{bosse2012} as a baseline method, which was first published in 2012, but has been optimized over the last ten years proving its robustness and accuracy under various robotic platforms and environments \cite{pavel2018,guo2019}.

\subsection{Simulation on Trajectory Optimization}

\begin{figure}[t]
\centering{
\subfloat[]{\includegraphics[height=.20\textheight]{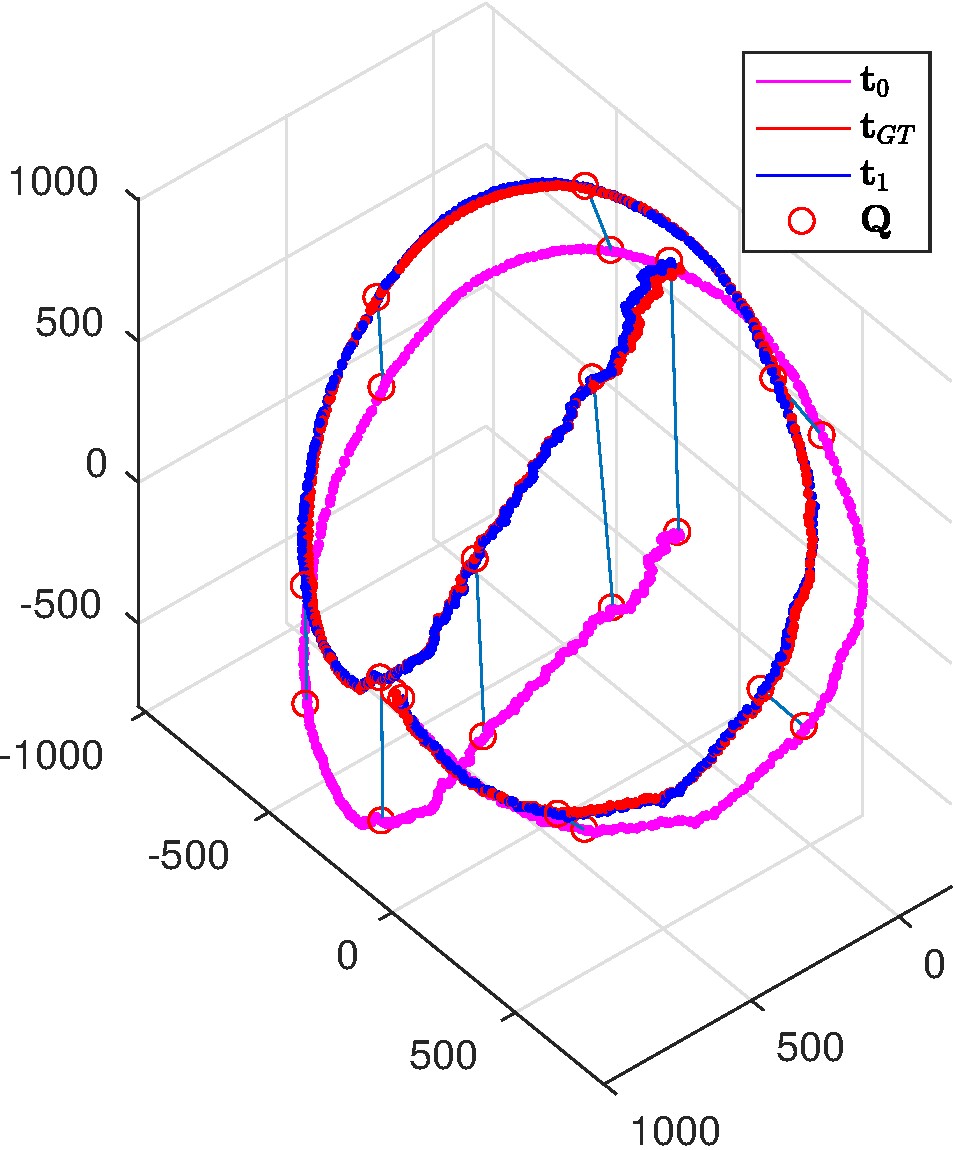}}\quad 
\subfloat[]{\includegraphics[height=.20\textheight]{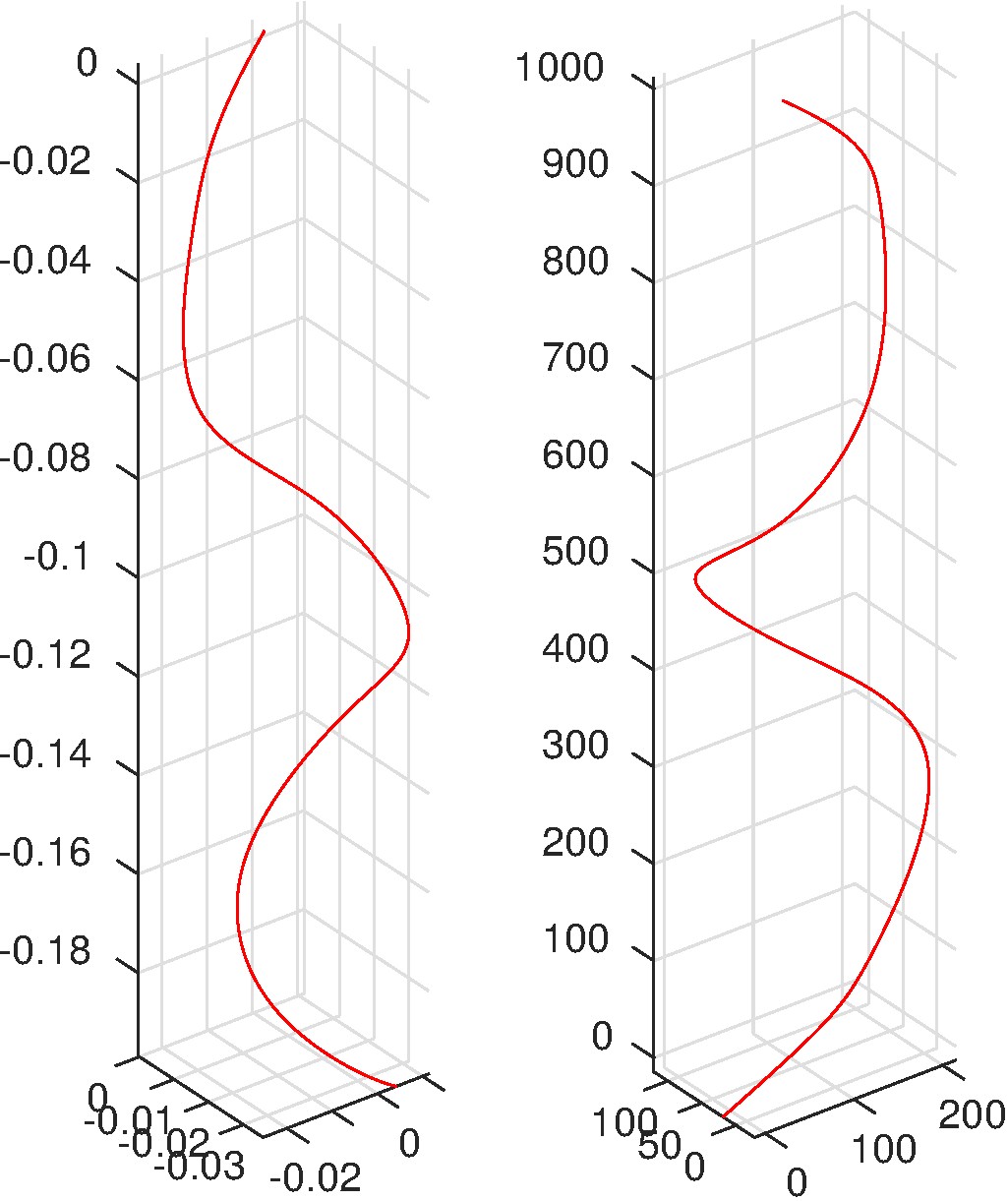}}
}
\caption{
(a) Visualization of the simulated trajectory (b)  its trajectory corrections (rotation left (radian), translation right (mm)). The initial trajectory $\textbf{t}_0$ (purple) is pulled by the corrections (red circles) towards the ground truth trajectory $\textbf{t}_{GT}$ (red). The correction in (a) visualize the correction in the first iteration of the optimization. The corrected trajectory $\textbf{t}_1$ (blue) went through several more iterations.   
}
\label{fig:correction_sim}       %
\end{figure}

We demonstrate the performance of the proposed method by comparing it to the most popular spline based trajectory optimization model through simulation in order to have quantitative analysis with the trajectory ground truth.

We have utilized a five second length of local window with a simulated angular velocity and linear acceleration at 100 Hz. The initial trajectory $\textbf{t}_0$ is built by an accumulation of the simulated angular velocity and linear acceleration with a bias and Gaussian noise. Randomly generated sparse surfel features (1000) are utilized for the trajectory optimization and assigned with a timestamp. For simplicity, we have only utilized the local surfel to fix the surfel registration constraint in Eq (\ref{eq:surfelmat_map}).
The simulated trajectory optimization and corrections are visualized in Figure \ref{fig:correction_sim}.

To demonstrate the advantage of the proposed method in local trajectory estimation, we present the comparison of the composition trajectory model and approximation model along with a few different parameters in Table \ref{tbl:simulated_results_CTSLAM}. Interpolation represents the interpolation method utilized for the cost function. The composition model in this experiment has 500 discrete trajectory poses $SE(3)$ where the interpolation at $\tau$ is by taking two poses around the time stamp and interpolating either on euclidean (Linear) or manifold (Linear $\mathfrak{s}\mathfrak{e}(3)$) space. Interpolation is directly calculated in the approximation model. "No. of States" is the actual dimension of the parameters in the optimization which is six times of the "No. of Compositions/Controls" as rotation vector and translation representation is utilized.
For the quantitative comparison, methods are evaluated by the absolute trajectory accuracy after the optimization.

\newcommand \figheight {.155}

\begin{figure*}[t]
\centering{
\subfloat[]{\includegraphics[height=\figheight\textheight]{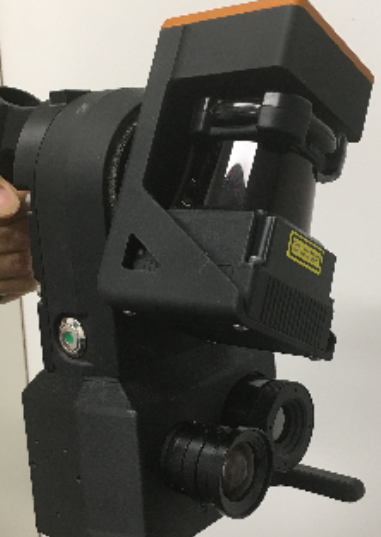}}\quad
\subfloat[]{\includegraphics[height=\figheight\textheight]{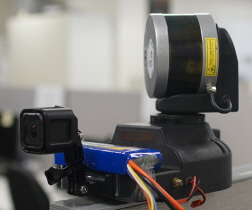}}\quad
\subfloat[]{\includegraphics[height=\figheight\textheight]{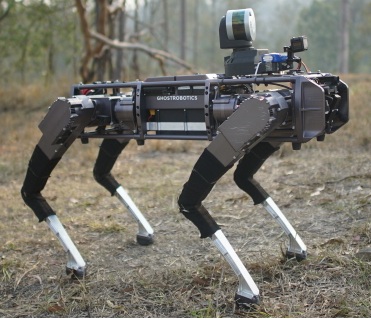}}\quad
\subfloat[]{\includegraphics[height=\figheight\textheight]{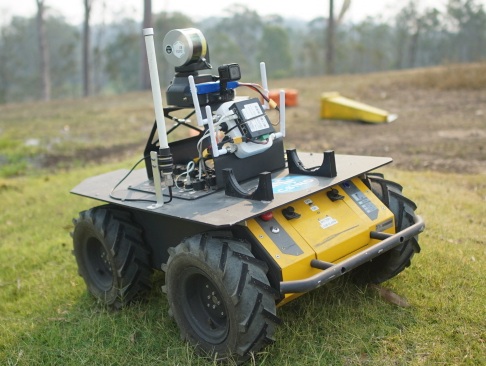}}\quad
}
\caption{ Experimental sensor payload configurations: (a) Hand-held single-beam LiDAR, (b) Hand-held multi-beam LiDAR, (c) legged robot with multi-beam LiDAR (d) wheeled robot with multi-beam LiDAR. 
}
\label{fig:sensors_robots}       %
\end{figure*}
\begin{figure*}[t]
\centering{
\includegraphics[width=.99\textwidth]{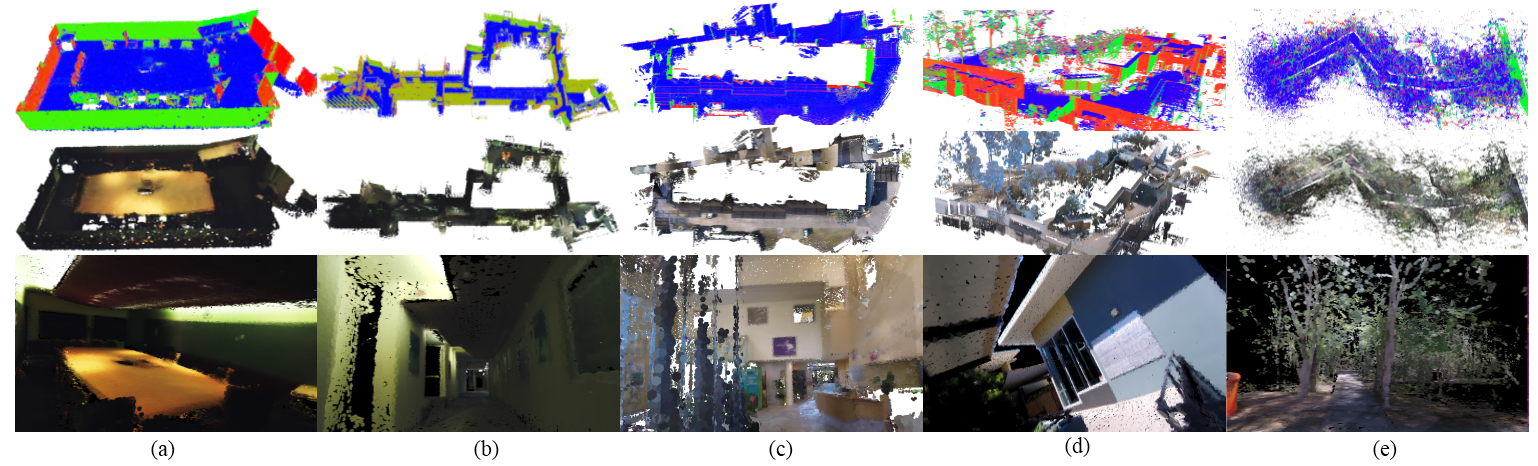}
}
\caption{Visualization of the datasets collected with the hand-held devices. {Starting from the top row, normal map, colorized map, details of the surfel rendered map.}
}
\label{fig:maps}       %
\end{figure*}

\begin{figure*}[t]
\centering{
\def\svgwidth{15mm}
\subfloat[]{\includegraphics[width=.45\textwidth]{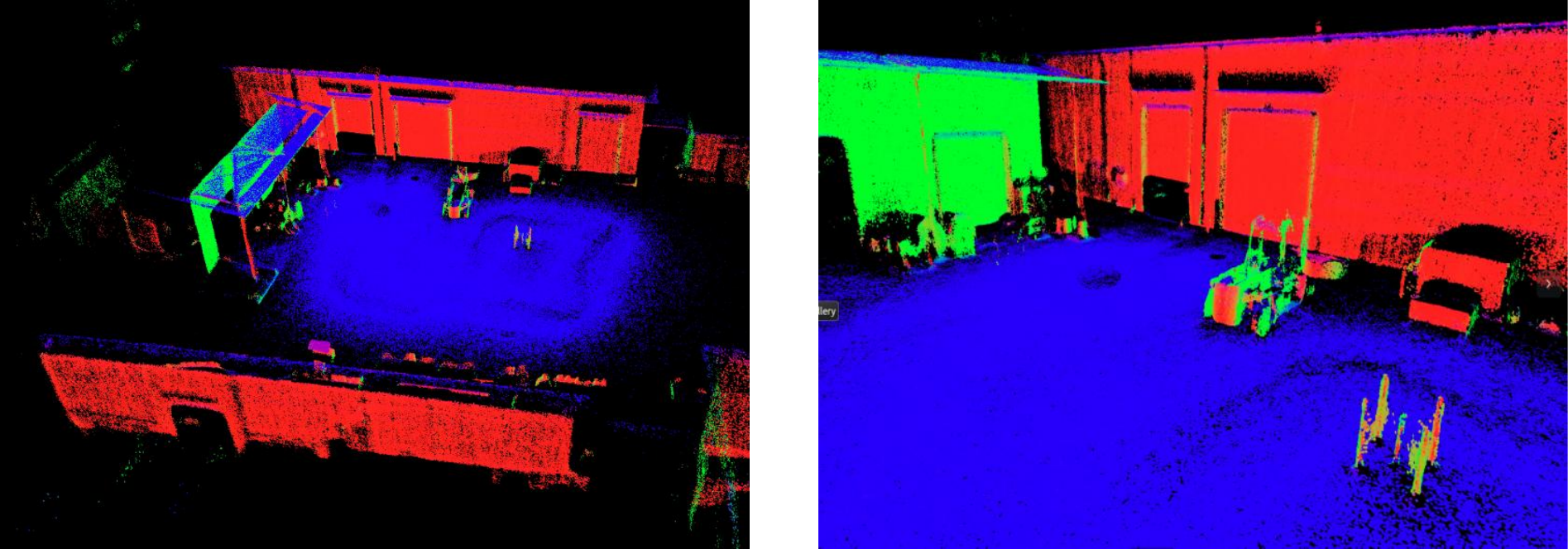}}\quad\quad
\subfloat[]{\includegraphics[width=.45\textwidth]{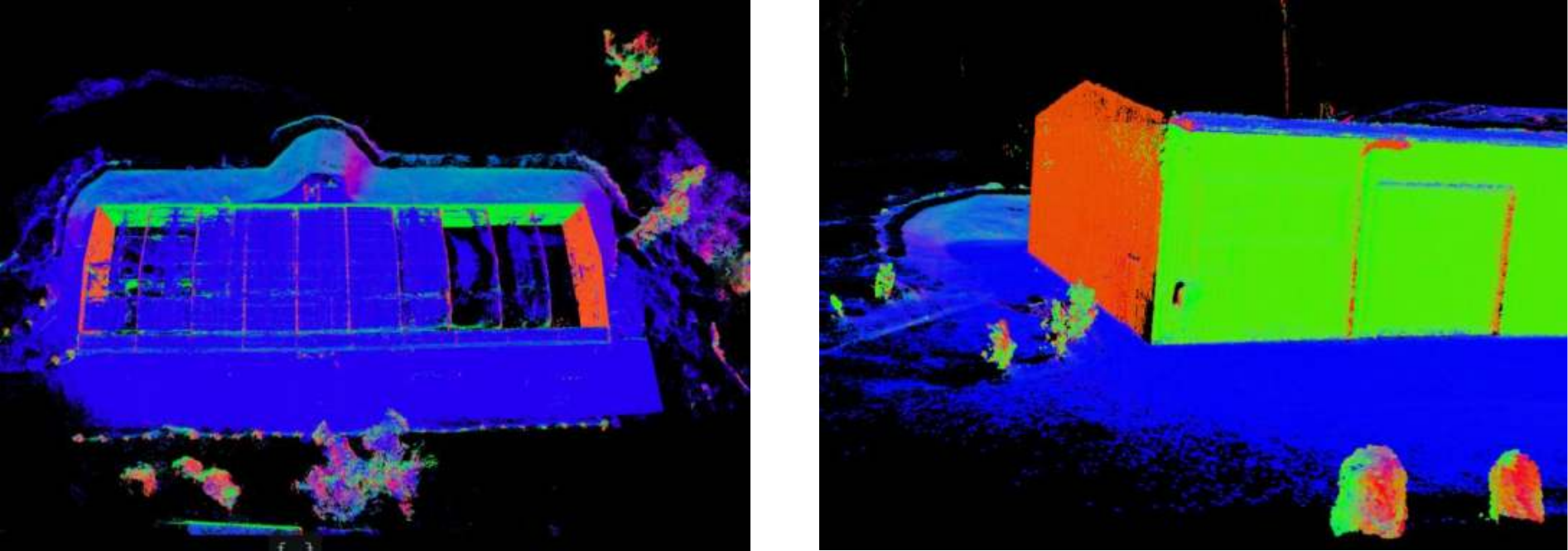}}
}
\caption{Visualization of the datasets collected from the robotic platforms. (a) and (b) are respectively collected by the legged robot and the wheeled robot around industrial areas. The duration and the size of the datasets are respectively (a) 2.5min, 72$\times$42(m) and (b) 7min, 38$\times$49(m).  
}
\label{fig:maps_robot}       %
\end{figure*}

\begin{figure}[t]
\centering{
\subfloat[]{\includegraphics[height=.150\textheight]{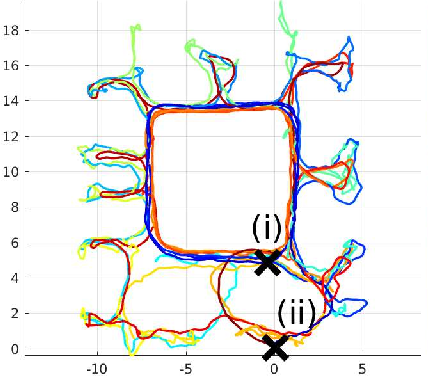}}
\subfloat[]{\includegraphics[height=.150\textheight]{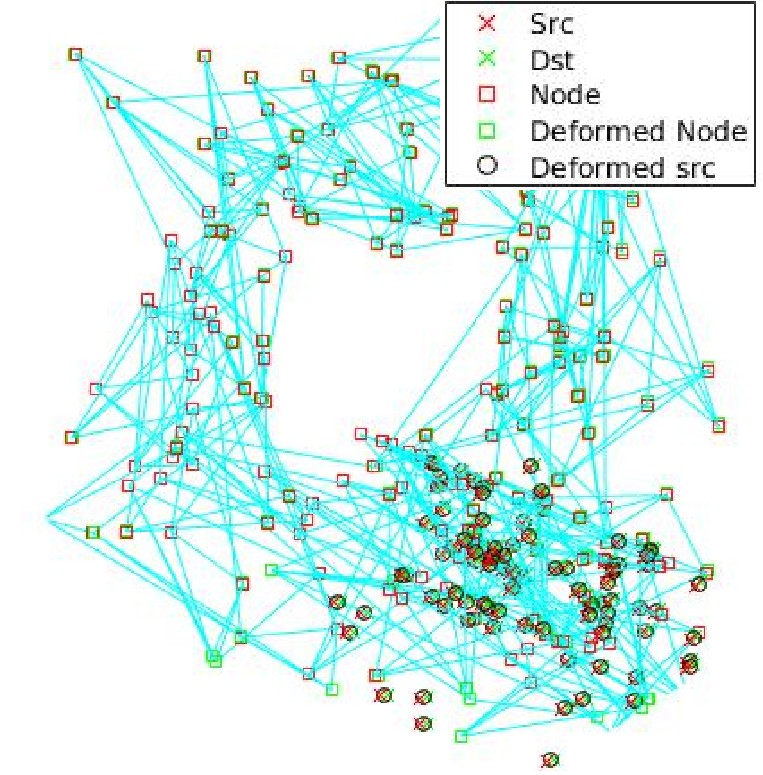}}

}
\caption{
Scanning trajectory and deformation graph of Figure. \ref{fig:mapdiff}. (a) Scanning path. {Our proposed} method {closes a loop} at (i) whereas trajectory based SLAM does at (ii). {The trajectory is colored by time: blue at the start transitioning to red at the end.} (b) Constructed graph {(red squares)} and its correction {(green squares)}. 
}
\label{fig:deformationgraph}       %
\end{figure}

The result in Table \ref{tbl:simulated_results_CTSLAM} shows that when the update is smoothed by spline and properly done on $SE(3)$ the Continuous-Time trajectory optimization result is better than the linear $SO(3) +\mathbb{R}^3$ composition method originally proposed in \cite{bosse2012} and the approximation model \cite{furgale2012}.

\subsection{Experimental Setup}

To demonstrate utility of the proposed method under various scenarios, we have collected multiple datasets with different sensor setups mounted to multiple robotics platform as in Figure \ref{fig:sensors_robots}. The collected datasets are categorized into three classes according to sensor types, environments, and robotic platforms.

Through Figure \ref{fig:maps} (a) to (c), the datasets are collected with a single beam hand-held 3D spinning LiDAR. The device consists of a spinning Hokuyo UTM-30LX laser, an encoder, a Microstrain 3DM-GX3 IMU, RGB camera. In the collected datasets sensors were moved with 0.9 $m/s$ and 0.7 $rad/s$ while rotor spins at 1 $rotation/s$. The datasets in Figure \ref{fig:maps} (d) to (e) are collected with a hand-held multi-beam LiDAR with Velodyne VLP-16, a Microstrain 3DM-GX3 IM and an independent Gopro camera. Gopro does not share a common clock with LiDAR.

The dataset in Figure \ref{fig:maps_robot} (a), (b) are presented to demonstrate the utility of the proposed method on realistic robotic platforms, respectively, a four legged robot and ground robot. For those dataset, the same sensor payload (Velodyne VLP-16 LiDAR, IMU and camera) was utilized while they are mounted on each platform as in Figure \ref{fig:sensors_robots}.

Also, we have collected datasets in different environments to demonstrate the robustness of the method against the different geometrical shapes. The datasets (a) and (b) in Figure \ref{fig:maps} are collected indoors whereas Figure \ref{fig:maps} (c) to (e) are collected outdoors. The outdoor dataset mixes structured (c) and (d) and unstructured places (e). %

\begin{figure}[t]
\centering{
\subfloat[]{\includegraphics[height=.145\textheight]{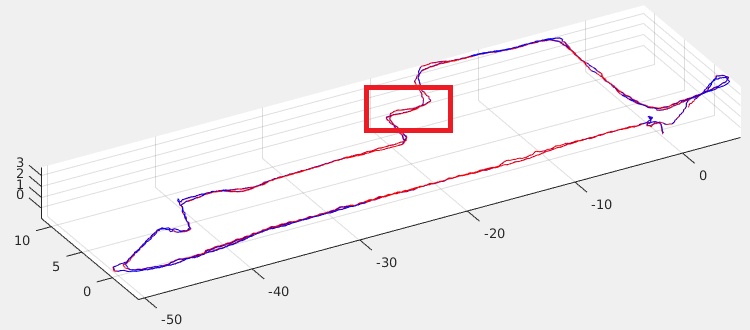}}\\
\subfloat[]{\includegraphics[width=.450\textwidth]{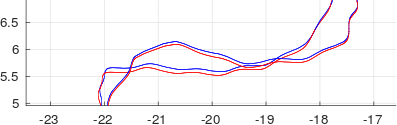}}
}
\caption{
Qualitative trajectory comparison between the global trajectory optimization (blue line) \cite{bosse2012} and the deformed trajectory (red line) of the map in Figure \ref{fig:maps} (c). {(b) details of the trajectory in the red box. Note that the trajectory includes two traverses. The unit is in meters.}
}
\label{fig:trajcomp}       %
\end{figure}

\begin{figure*}[t]
\centering{
\includegraphics[width=.99\textwidth]{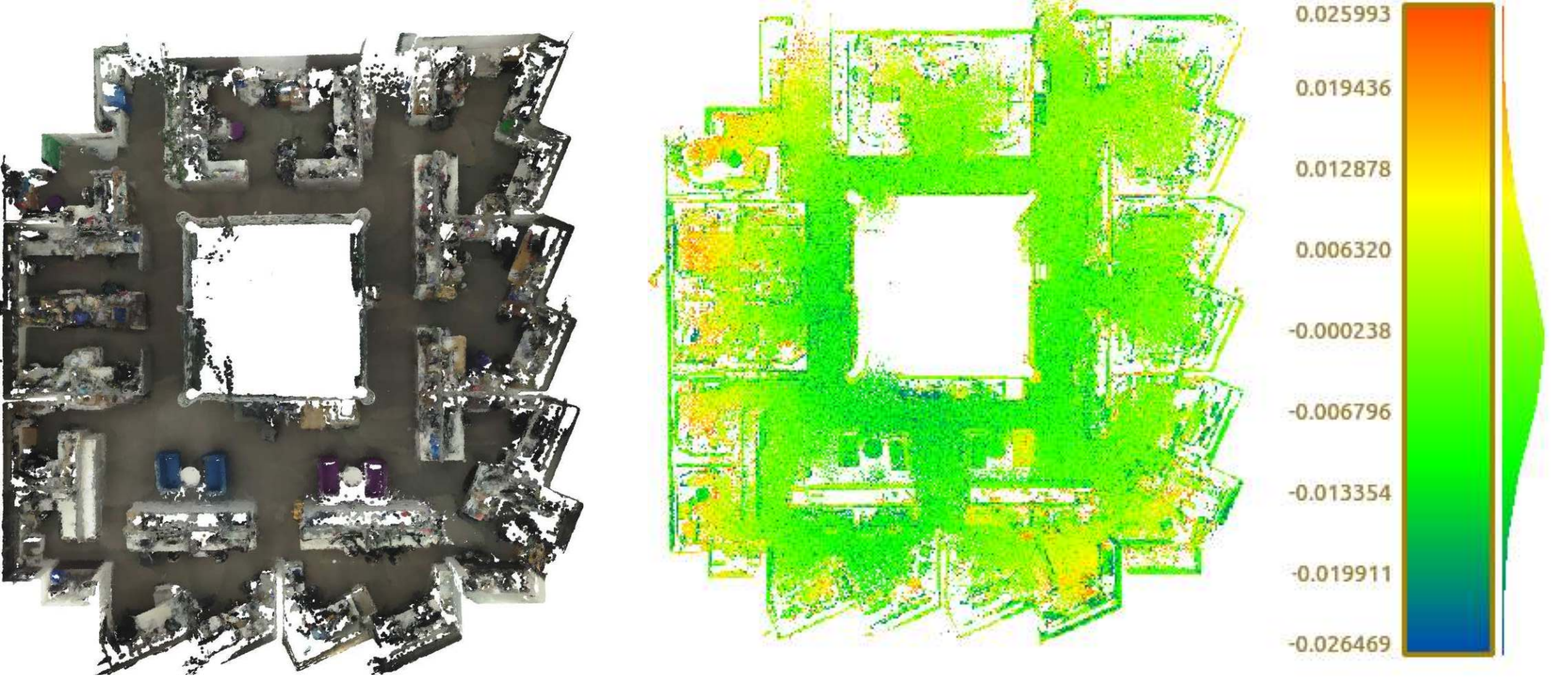}
}
\caption{
Visualization of the dataset (f): colorization and structural difference. Colour in the right side figure represent the difference between the deformed map and the baseline map by \cite{bosse2012}. Bar on the right represents the histogram and range of the difference in color. The error range on the bar is in meter scale. 
}
\label{fig:mapdiff}       %
\end{figure*}

Spatiotemporal extrinsic parameters between the LiDAR and the rest of the sensors (camera, IMU) are estimated using the method in our previous publication \cite{park2020,vidas2013,moghadam2013}. The spatiotemporal parameters are utilized for a colorization of surfel. 
Shader-based surfel render is developed for the colorization and 3D scene rendering. Some of the scenes of the dataset are rendered and presented in the third row of Figure \ref{fig:maps}.    
For the colorization, a distorted image is directly utilized along with the calibration parameters in the shader to avoid loosing information and reduce computational cost. %
Some of the datasets require a global loop closure detection. The global loop closure is found by 3D and 2D combined loop closure detection method \cite{park2019,guo2019} and the misalignment is estimated using the sequential method described in Section \ref{sec:deformation} using only 3D point cloud. Detected loops are closed by the deformation graph as depicted in Figure \ref{fig:deformationgraph} (b). %

\subsection{Trajectory and Structure Comparison}

To evaluate the trajectory estimation accuracy of the proposed method, we compared the deformed trajectory with the globally optimized trajectory from the benchmark method~\cite{bosse2012}. Note that in the proposed method, the trajectory is not required to be stored. They are saved and deformed only to evaluate accuracy of the proposed method. For the comparison, we utilized the absolute trajectory Root Mean Square Error (RMSE) metric, which estimates the Euclidean distances between the deformed trajectory and the globally optimized trajectory. %

Table \ref{tbl:trajestimation} shows the error and trajectory statistics of each dataset visualized in Figure \ref{fig:maps}. The trajectory error of each estimations varies from 0.2 to 0.5 meters with high correlation to the map size. The trends shows that the difference gets larger when the map scale is growing, which is reasonable considering that the proposed method does not count gravity direction in the deformation graph optimization. The initial position is aligned with the gravity direction but when the sensor scans further area from the initial location the gravity alignment faints, making the difference. A qualitative comparison of trajectories from the dataset (b) is presented in Figure \ref{fig:trajcomp}. The visual comparison shows that local details are well preserved and similar to the globally optimized trajectory, but there were overall 0.1 meters offset between the two trajectories.

Also, to visualize the structural difference of highly redundant scanning case, we registered the two point clouds respectively from the baseline method \cite{bosse2012} and the proposed method and calculated the point-to-mesh distances in Figure \ref{fig:mapdiff}. For the point-to-mesh distance, a fine mesh is extracted from the globally optimized point cloud of the baseline method \cite{bosse2012}. The result shows the reconstruction difference was within $\pm$0.02 meters but it was growing on the outer side.

\begin{table*}[t]
\centering

\begin{tabular}{lcccccccc}
\hline\noalign{\smallskip}
 Dataset & \multicolumn{1}{c}{(a) Small room}  & \multicolumn{1}{c}{(b) Multiple Floors}  & \multicolumn{1}{c}{(c) In/outdoor }   & \multicolumn{1}{c}{(d) Outdoor}  & \multicolumn{1}{c}{(e) Unstructured} & \multicolumn{1}{c}{(f) Office } \\
\hline\noalign{\smallskip}
Traj Error(m)      &0.173  &0.245    &0.233  &   0.552    &  0.301     &0.189 \\
Length(m)       &130    &300    &360    &210    &110    & 330 \\
Time(min)       &6.1    &11.4   &9.1    &  3.5     &4.1     &14.6 \\
Size(m)         &10$\times$6    &55$\times$20   &60$\times$25  &60$\times$69 &47$\times$17 & 20$\times$20\\
No. of Beams    &Single &Single &Single &Multi  &Multi &Single \\

\noalign{\smallskip}\hline\noalign{\smallskip}   

\end{tabular}
\caption{Absolute trajectory estimation {RMSE in meter between the deformed trajectory and the globally optimized trajectory (CT-SLAM \cite{bosse2012}).}}
\label{tbl:trajestimation}

\end{table*}

\subsection{Surface Estimation}

\begin{figure*}[h]

\begin{center}

\subfloat[]{\includegraphics[height=5.3cm]{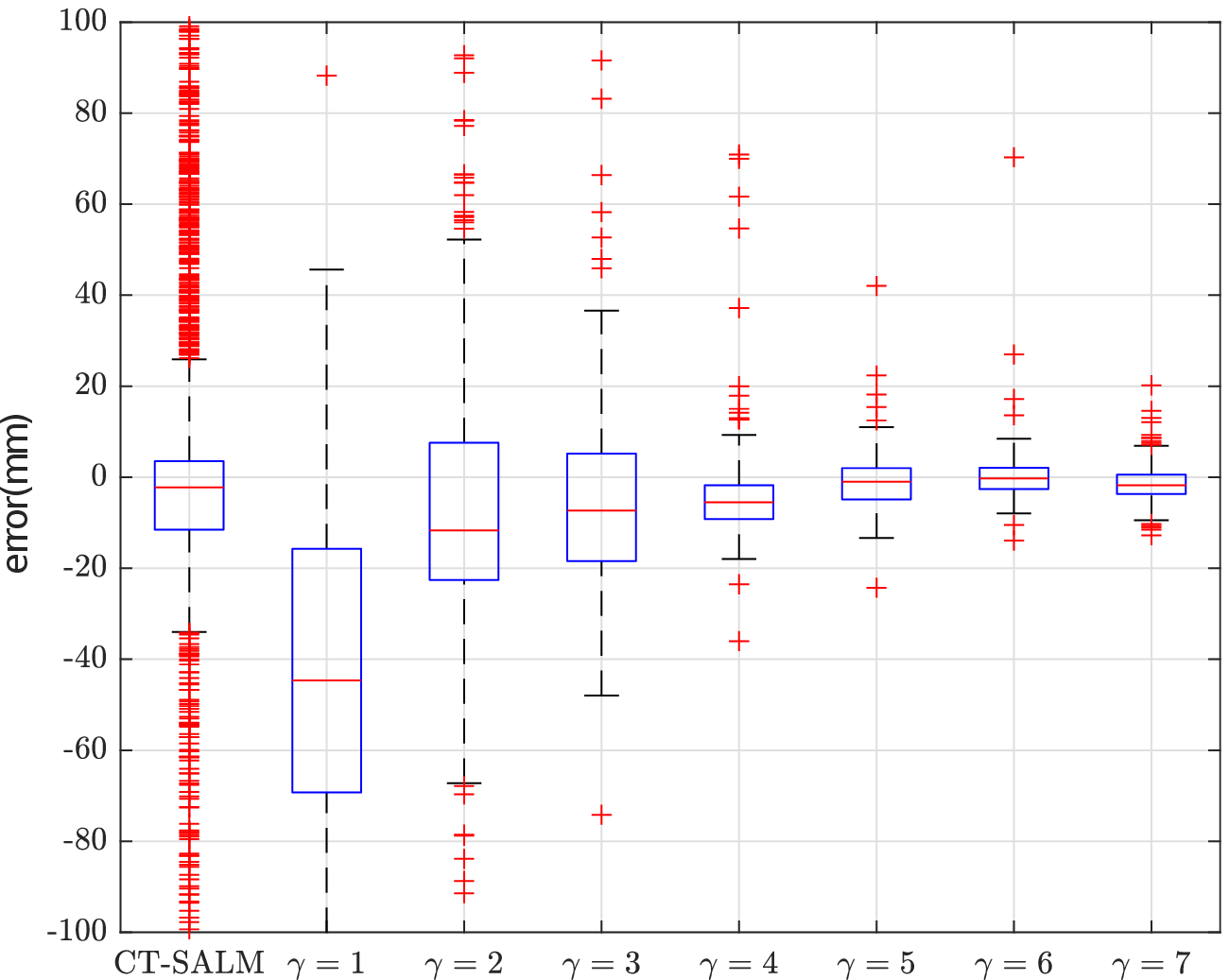}}%
\quad
\subfloat[]{\includegraphics[height=5.3cm]{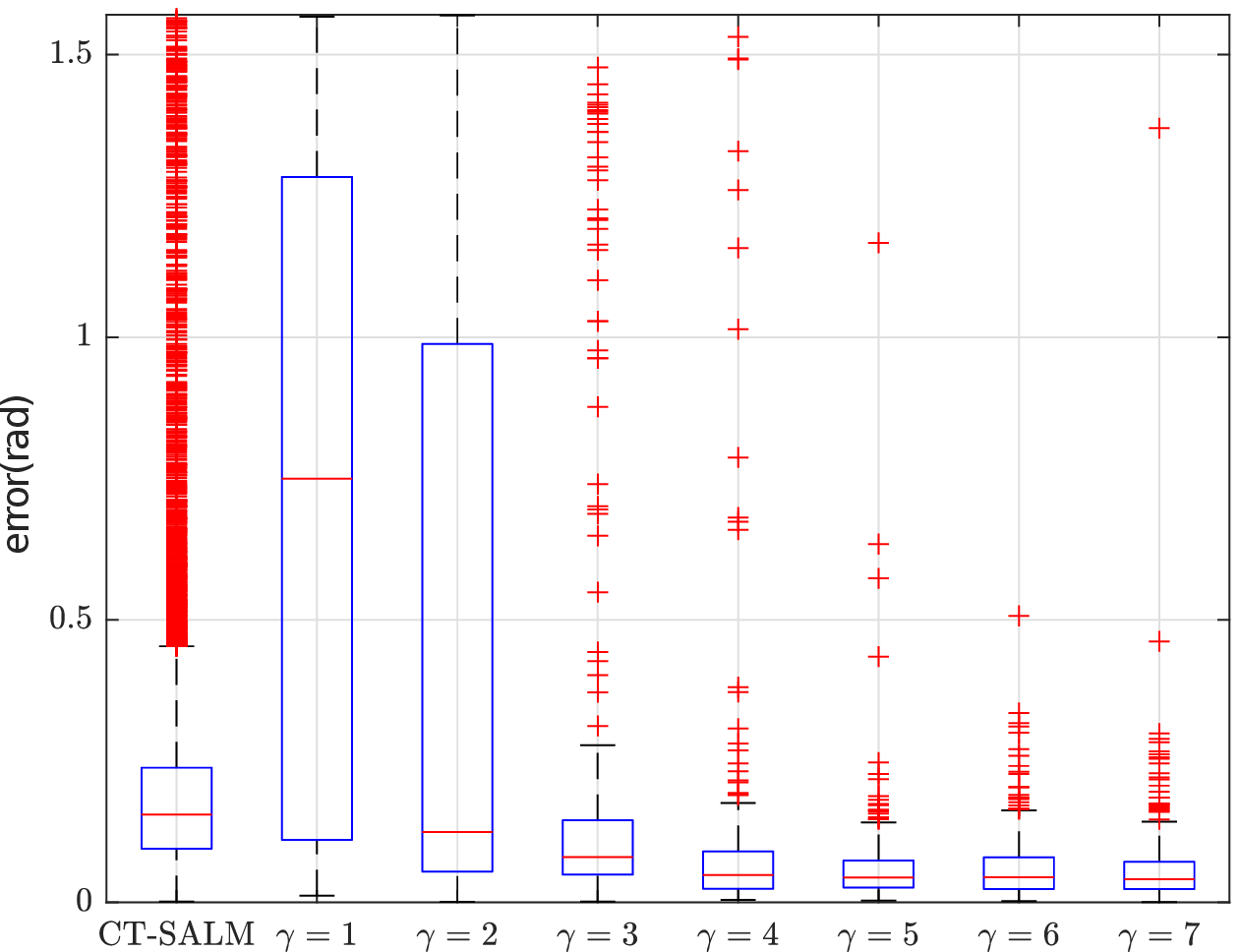}}%
\caption{ Number of fusion versus accuracy of a planner patch((a) centroid (b) normal).    
}
\label{fig:patch_analyisis_gam}
\end{center}
\end{figure*}

\begin{figure*}
\centering{
\def\svgwidth{150mm}
\includegraphics[width=.80\textwidth]{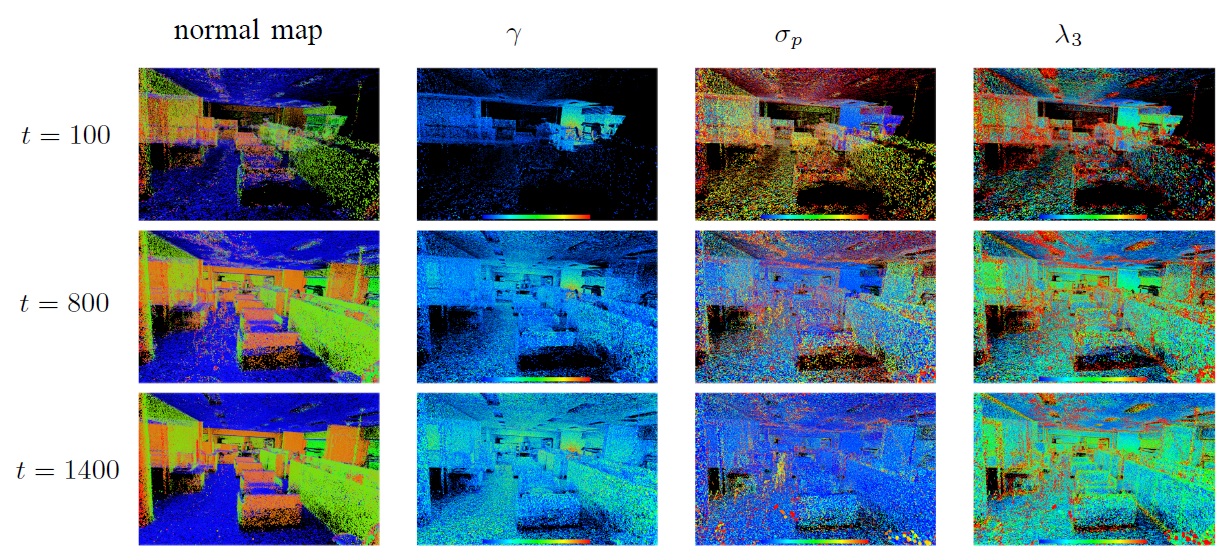}
}
\caption{Map building process over time. Color represents attribute values of each surfel. Colors shows respectively normal direction in the first column, number of observation (blue means low, red means high) in the second column, normal and position uncertainty (blue means low, red means high) in the third and forth column.
$t$ is in second.}
\label{fig:progress_ink}
\end{figure*}

For a quantitative comparison on real datasets where ground truth is not available, known planar surfaces such as a floors or walls are utilized for calculating the relative noise level. 

Table \ref{tbl:surfaceestimation} shows statistical analysis of the multiple planar areas. The error in the table represents the mean projective distance, which is calculated from the mean plane of each patch. 
With the same data extracted from the planar patches, the analysis result on the relationship between the number of fusion steps and accuracy is given in Figure \ref{fig:patch_analyisis_gam} along with the comparison with the original raw points cloud on the left most side of each figure. For relative comparison, the distance of each points in the area from the average plane of the area were used to calculate normal direction errors and projective distance errors. $\gamma$
represents the number of fusion where surfels with the same number of fusions
are utilized for the statistics. For example, for the calculation of $\gamma = 2$ surfels that are fused only 2 times are utilized. 
The result in Figure \ref{fig:patch_analyisis_gam} indicates that the errors drastically are reduced when surfels are fused more than four times reaching $\pm$10mm and 0.2rad ($\approx$5deg) error range. The surfels with low numbers of observations are removed from the global map after some period of time as they tend to have higher possibility of being outlier or mixed pixels \cite{tuley2005}. 
The development of the map and the surfel statistics changes over time are visualized in Figure \ref{fig:progress_ink} with the dataset (f). The figure shows how the number of surfel observations impact uncertainties when the sensor makes multiple traverses on the same place.

\begin{table*}[]
\centering

\begin{tabular}{crrrrrrrr}
\hline\noalign{\smallskip}

&  \multicolumn{4}{c}{CT-SLAM \cite{bosse2012}}  &  \multicolumn{4}{c}{Proposed method} \\ \cline{2-9}
\multirow{2}{*}{Patch No.}&  \multicolumn{2}{c}{Position Err.} & \multicolumn{2}{c}{Normal Err.} &  \multicolumn{2}{c}{Position Err.} & \multicolumn{2}{c}{Normal Err.} \\ \cline{2-9}
&       mean  & std. & mean & std. & mean  & std. & mean & std.\\ 
\hline
\noalign{\smallskip}%
a   &  8.2 & 14.8  &0.101  & 0.082   & \textbf{3.4} &\textbf{4.7}  &  \textbf{0.069}  & \textbf{0.075}   \\
b   & 9.3 & 16.8  &  0.120   & 0.116    & \textbf{3.2}& \textbf{4.4} &\textbf{0.085}  &  \textbf{0.076} \\
c   & 8.9  & 17.3 &   0.092  &  0.120   &  \textbf{3.4}&   \textbf{6.2}  &\textbf{0.076}&    \textbf{0.082} \\
d   & 9.4 &  17.0 & 0.085  &   0.099   &\textbf{3.9} &   \textbf{5.3} & \textbf{0.078}&   \textbf{0.080}  \\
f   & 8.0  & 13.7  & 0.096 &   0.087  &  \textbf{3.4} &   \textbf{4.7} & \textbf{0.083}&   \textbf{0.080}   \\
g       & 9.1  & 16.0  & 0.106 &    0.115  &  \textbf{4.5} &  \textbf{6.5} &  \textbf{0.078}&  \textbf{0.082}  \\
\noalign{\smallskip}\hline\noalign{\smallskip}   
\end{tabular}
\caption{Surface estimation statistics. 
The projective distance errors and the normal error. Note that the point cloud from CT-SLAM \cite{bosse2012} is undistorted by the globally optimized trajectory but unfused raw points. Units are mm for position error and radian for normal direction.
}
\label{tbl:surfaceestimation}
\end{table*}

\subsection{Evaluation of Localization for Loop Closure}

For the initial loop closure detection, we have utilized the visual place voting method \cite{bosse2013}. 
Once a place recognition triggers the proposed localization for
the loop closure, a procedure for estimating 6 DoF misalignment starts where
it finds only single misalignment between the very first revisited place found from the previous map and the new map. In the sense that the process includes 6 DoF registration with unknown initial guess, the problem is closer to a large scale global registration problem \cite{yang2016} or kidnap recovery
problem rather than a place recognition problem \cite{lynen2014}.
To this end, we compare our method with the well known global ICP registration libraries \cite{zhou2018,tombari2010Daniilidis} with different initial guesses.  

We have utilized outdoor and indoor mixed point cloud datasets to evaluate the localization performance. The globally optimized trajectory was utilized for calculating the ground truth.
To estimate the robustness against initial guess, initial poses are randomly generated with ten different loop closure scenarios. We have set the covariance for the random misalignment generation to be larger along the z axis (gravity direction) and smaller in x and y axis (orthogonal to gravity direction). The 6 DoF misalignment is estimated 50 times each with ten different locations which simulates 500 loop closure triggers. For each loop closure trigger, misalignment starts sequential estimation over different places along the trajectory which exactly simulate the loop closure procedure with real data. 

The RMSE of the estimated pose is listed in Table \ref{tbl:noise} along with sparse ICP algorithm (a), and the proposed method (e) and the other global registration algorithms (b), (c). In (b), Open3D \cite{zhou2018} global registration is utilized with an initialization by FPFH \cite{rusu2009} feature and RANSAC.
Similar to (b), SHOT was utilized for the initial pose estimation, then refined by point-to-plane ICP in (c). Longer localization processing time drastically increases map fusion cost between the active map and the global map. Hence, we are not comparing to the time consuming algorithms such as GO-ICP \cite{yang2016} which could take several minutes to hours to process a single pair of point cloud.

The experiment shows that the surfel ICP (a) is prone to a local minima problem with a large displacement. The large displacement frequently occurs especially in a large scale mapping scenarios. The pose estimation from Open3D registrations (b) is generally reasonable regardless of the initial guess. However, occasional registration failures occurred due to the failure in finding feature matching.
SHOT (c) showed a moderate rotation estimation, yet the estimated translation often suffered from outliers even with a close initial guess. This is assumed to be due to the failure in the feature based pose initialization. The proposed method (d) shows the most stable result. Although, both translation and rotation estimation is slightly increasing reaching up to 6 cm and 0.004 rad ($\approx$0.22 deg) in the hard case scenario but they are within a reasonable range and still better than the other algorithms.

\begin{table*}[t]
\centering
\begin{tabular}{lcccccccc}
\hline\noalign{\smallskip}
Initial  & \multicolumn{2}{c}{(a) Sparse ICP }  &   \multicolumn{2}{c}{(b) Open3D \cite{zhou2018}}  & \multicolumn{2}{c}{(c) SHOT \cite{tombari2010Daniilidis} } & \multicolumn{2}{c}{(d) Proposed method} \\
 \cline{2-9}
Guess&       ${e_\textbf t}$  &  ${e_\textbf r}$ &  ${e_\textbf t}$  &  ${e_\textbf r}$ &  ${e_\textbf t}$  &  ${e_\textbf r}$ &  ${e_\textbf t}$  &  ${e_\textbf r}$ \\ 
\hline
\noalign{\smallskip}%

Easy    &  0.04(0.04)&  0.01(0.01)&    0.30(0.65)&  0.03(0.06)&1.49(1.52)&    0.07(0.12)                   &  \textbf{0.03(0.01)} &  \textbf{{0.001}(0.0005)}\\
Medium  &  0.40(0.51)&  0.19(0.31)&   1.64(2.39)&  0.30(0.38)   &1.53(1.55)&    0.11(0.29)            &  \textbf{{0.04}(0.02)}&  \textbf{{0.004}(0.001)}\\
Hard    &  2.52(0.87)&  2.42(1.55)&   13.8(22.4)&  0.38(0.63)    &7.59(18.56)&    0.30(0.71)               &  \textbf{{0.06}(0.04)}&   \textbf{{0.004}(0.001)}\\ 
\noalign{\smallskip}\hline\noalign{\smallskip}   
\end{tabular}
\caption{Localization accuracy comparison. The estimations are compared to the ground truth to calculate error norm of translation ${e_\textbf t}$ and rotation ${e_\textbf r}$ with standard deviation in parentheses. Units are rotation vector norm and meter. For each noise level, initial poses are randomly generated according to the following parameters:   Easy $\sigma_{\theta z=10},\sigma_{\theta
xy=1},\sigma_{t=0.5}$, Medium $\sigma_{\theta z=50},\sigma_{\theta
xy=5},\sigma_{t=5}$, Hard $\sigma_{\theta z=100},\sigma_{\theta
xy=20},\sigma_{t=50}$ ($\sigma_{\theta}$ in deg, $\sigma_{t}$ in meter). %
}
\label{tbl:noise}
\end{table*}

\section{Discussion}
\label{sec:discussion}

\subsection{Trajectory Optimization}
The composition model stores 500 raw poses and only optimizes the trajectory corrections whereas the approximation model entirely represents the trajectory as the abstracted few spline control points (Figure \ref{fig:correction_sim}). Because of this difference, with the same state dimension to be optimized, the approximation method has significantly lower accuracy.
With relatively higher motion resolution (101 control points), the approximation method follows the ground truth trajectory reasonably but high frequency motions are ignored. The ignorance of the high frequency motion propagates noise to the map when the scanning equipment moves in a high frequency motion such as in the case where the device is mounted on a wheeled platform driving outdoor. Considering LiDAR measurements reach up to several hundred meters, even a small angular difference can cause a large displacement of the points.

\subsection{Map Estimation}

While the local details are very similar to the baseline method in Figure \ref{fig:trajcomp}, there were overall structure movement about 0.1 meter which make similar movement in the deformed trajectory as shown in Figure \ref{fig:trajcomp} (b). This could be problematic when the scanning scenario includes partial revisit of a space which does not make enough overlap to trigger a local loop closure. While the map-centric approach solves the time complexity problem, this is fundamental weakness of the proposed method when used with a long range LiDAR. Thus, we suggest to reduce the fusion range.

The highly redundant scanning case in Figure \ref{fig:mapdiff} showed slightly different pattern to the dataset (b). The center part is close to the baseline method but outer part of the map showed difference of $\pm$ 0.02 meters. This is due to the loop closure occur in a the first traverse as visualized at (i) in Figure \ref{fig:deformationgraph} (a). The short distance make relatively small misalignment at the loop closure therefore the deformed map around this part is similar to the baseline method. However, when a new area is explored and the map grows, the difference starts to increase. Note that the difference in this case is not because of the map deformation but because of the accumulated small drifts.

Statistics in Table \ref{tbl:surfaceestimation} shows that the estimated surface is up to three times less noisy than the point cloud generated using baseline CT-SLAM~\cite{bosse2012}. 
However, repeated non-Gaussian noises such as noises from mixed pixel problems \cite{tuley2005} are not effectively fused by the proposed method. The repeated non-Gaussian noise pattern usually appears from stationary scans.

\begin{figure}[t]
\centering{
\includegraphics[width=.480\textwidth]{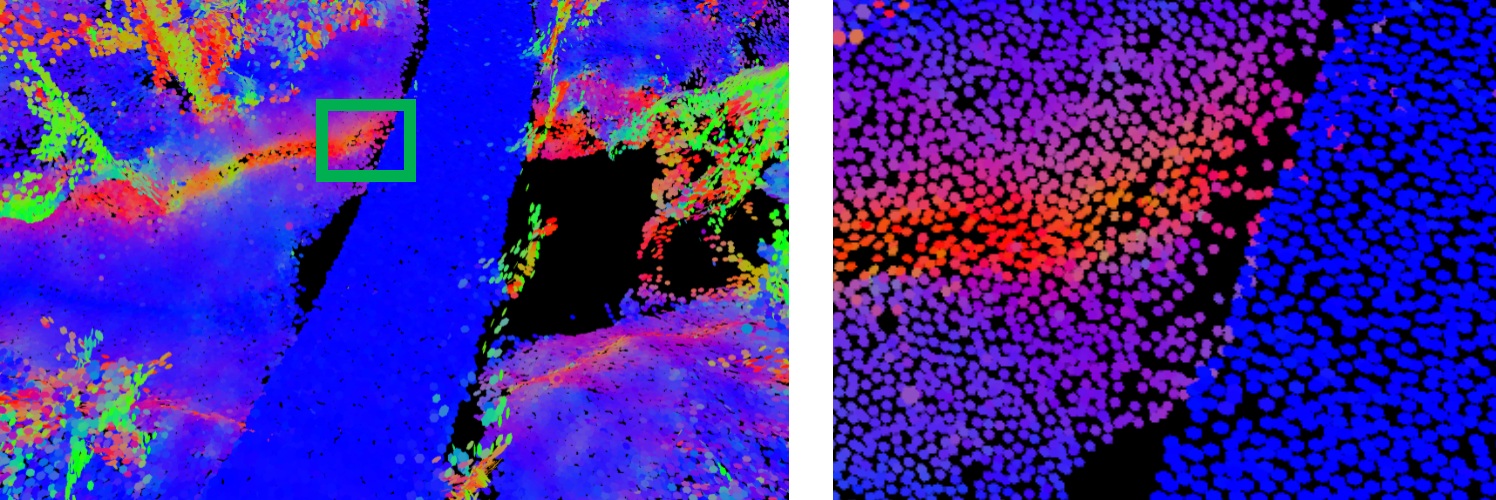}
}
\caption{ Visualization of surface resolution preservative fusion. Note that the surfel size is reduced in the left figure to show the surfels that are forming the surface. The color represents normal direction.}
\label{fig:mapdetails}       %
\end{figure}

Regardless of the number of the original raw points in the proposed method, the surfel density in patches are uniformly maintained after fusion. Figure \ref{fig:mapdetails} visualizes a small patch of the dataset (e) with a reduced surfel size to show that the surface forming surfels are non-redundant and uniformly distributed. However, because of the restriction in surface resolution, objects which are smaller than the given surface resolution are often ignored, losing some details of scene. When, an excessive surface resolution is utilized computational cost significantly increases due to the matching and fusion cost and also without increasing the reconstruction quality. With our experimental sensing payloads, we found that 0.02 meters or higher resolution were acceptable.

The result in Figure \ref{fig:progress_ink} shows that after the first traverse of the map ($t$=100 sec),  the number of observations of each surfel increases (second column) while the uncertainties decrease (third and forth column). The uncertainty of the overall surfel positions decrease whereas the scatterness along the z direction of the surfels around edges does not decrease after some point. This is due to the way the scatter matrix is extracted from the point cloud where surfels around edges and corners always have a high eigenvalue in the z direction. 

\subsection{Localization for Loop Closure}

Compared to the methods (a), (b), (c) in Table \ref{tbl:noise} our proposed sequential metric localization method utilizes more point clouds collected at different locations. However, rather than utilizing a single large chunk of the point cloud, the proposed method (d) segments the map into frame level and runs independent registrations. While this approach provides an evaluation metric to decide the reliability of the current estimation, it also offers cross-validation ability between the independent misalignment estimations. These two components in the proposed method significantly contributed in reducing localization failures. However, the translation error is still relatively higher than the map reconstruction accuracy. Considering the surface reconstruction quality is often goes below 1 centimeter, it is desirable to reduce the metric localization accuracy to the similar level.

\section{Conclusion}
\label{sec:conclusion}

We presented a new approach for dense LiDAR-based map-centric SLAM. Our main contribution in this paper is the identification of the problems that occur when applying map-centric dense mapping approach to LiDAR sensors, followed by a new SLAM framework that successfully adapted the beneficial characteristics of the map-centric approach and continuous-time representation. 

We have tested our method with 8 dataset collected under various challenging places in different scales with 4 different sensor types and configurations. Our experimental results indicate that, the deformed map showed only 2 centimeters difference, while the surface estimation is up to 3 times less noisy after multiple revisits. Also, the sequential misalignment method gives a stable and accurate localization estimation within 6 centimeters even with the hardest cases where other common methods fail. %

While we saw the possibility of LiDAR device for map-centric SLAM, we found couple of limitations such as growing map distortion or the partial observation problem. The distortion in the 60 meter scale map reached to 10 centimeter and expected to be larger with extended map size. However, we believe these weakness can be solved by further introducing additional priors and constraints. So, we leave these problems as future works.

\appendices

\section{Surfel fusion}
Here we briefly describe the method that we utilise for fusion of surfel ellipsoids. If points were observed noise-free, i.e., LiDAR points were distributed according to $\mathcal{N}\{\boldsymbol{\mu},\textbf{X}\}$ where $\boldsymbol{\mu}$ and $\textbf{X}$ are the true surfel parameters, then a normal inverse Wishart model may be used to estimate the implied ellipsoid by accumulating the overall mean and sample covariance. A fast, approximate method for accommodating noisy observations of the form $\mathcal{N}\{\boldsymbol{\mu},\textbf{X} + \textbf{Q}\}$ is proposed in \cite{FelFra11}. The method maintains a state which includes the estimate of the surfel centroid $\hat{\boldsymbol{\mu}}$, the covariance of this estimate, $\hat{\boldsymbol{\Sigma}}$, the accrued sample matrix for the extent, $\hat{\boldsymbol{\Xi}}$, and the count of points included in the estimates, $\nu$. The updates due to a new surfel with $n$ points with mean $\bar{\mathbf{z}}$, scatter matrix $\bar{\textbf{Z}}$ and measurement noise $\textbf{Q}_w$ are:
\begin{align}
\hat{\boldsymbol{\mu}}^+ &= \hat{\boldsymbol{\mu}} + \textbf{K}(\bar{\mathbf{z}} - \hat{\boldsymbol{\mu}}) \\
\hat{\boldsymbol{\Sigma}}^+ &= \hat{\boldsymbol{\Sigma}} - \textbf{K}\hat{\boldsymbol{\Sigma}} \\
\textbf{K} &= \hat{\boldsymbol{\Sigma}} \textbf{S}^{-1} \\
\textbf{S} &= \hat{\boldsymbol{\Sigma}} + \frac{\textbf{Y}}{n} \\
\textbf{Y} &= \hat{\textbf{X}} + \textbf{Q}_w \\
\hat{\textbf{X}} &= \frac{\hat{\boldsymbol{\Xi}}}{\nu - n_z - 1}
\end{align}
where $n_z$ is the measurement dimension, an the extent update of:
\begin{align}
\nu^+ &= \nu + n \\
\hat{\boldsymbol{\Xi}}^+ &= \hat{\boldsymbol{\Xi}} + \bar{\textbf{N}} + \bar{\textbf{Y}} \\
\bar{\textbf{N}} &= \hat{\textbf{X}}^{1/2}\textbf{S}^{-1/2} \textbf{N} \textbf{S}^{-T/2} \hat{\textbf{X}}^{T/2} \\
\bar{\textbf{Y}} &= \hat{\textbf{X}}^{1/2}\textbf{Y}^{-1/2} \bar{\textbf{Z}} \textbf{Y}^{-T/2} \hat{\textbf{X}}^{T/2} \\
\textbf{N} &= (\bar{\mathbf{z}}-\boldsymbol{\hat{\mu}})(\bar{\mathbf{z}}-\hat{\boldsymbol{\mu}})^T
\end{align}
The estimate of the covariance defining the object ellipsoid is $\hat{\textbf{X}}$. An alternative method using variational inference has also been proposed \cite{Org12}, but this requires iterative solution, and is avoided for performance reasons.

\section{Sequential Pose Fusion}

The Bayesian fusion provides a closed-form solution on the vector fusion problem \cite{park2017c}. However, directly applying the Euclidean Bayesian fusion on the poses causes a convergence to a suboptimal as the pose vector is on manifold. Thus, we utilize the sequential {$SE(3)$} pose fusion approach proposed in \cite{barfoot2014,park2019}.

This begins by modeling an error $\boldsymbol{\epsilon}_n \sim\ \mathcal{N}(\textbf{0},\boldsymbol{\Sigma}_n)$ between an individual estimation of a pose $\textbf{T}_n$ and the ground truth $\textbf{T}_{GT}$ as 
\begin{equation}
\begin{split}
\boldsymbol{\epsilon}_n & = {\log}(\textbf{T}_{GT}\textbf{T}_n^{-1})\\
& = {\log}(\textbf{e}^{[\boldsymbol{\xi}]_{\times}}\textbf{T}^*\textbf{T}_n^{-1})\\
& = {\log}(\textbf{e}^{[\boldsymbol{\xi}]_{\times}}\textbf{e}^{[\boldsymbol{\xi}_n]_{\times}})
\end{split}
\label{eq:TT}
\end{equation}
where the second and third equation substitute the ground truth which is not directly available and convert the error as a function of a small perturbation $\boldsymbol{\xi}\in\mathfrak{s}\mathfrak{e}(3)$ regarding our best guess so far on the pose $\textbf{T}^*$ and the individual estimation. Thus, given $N$ poses the best pose that minimize the sum of the error can be achieved by iteratively minimizing the following cost function regarding $\boldsymbol{\xi}$.
\begin{equation}
V = \frac{1}{2}\sum_{n}^{N}\boldsymbol{\epsilon}_n^\top\boldsymbol{\Sigma}_n\boldsymbol{\epsilon}_n
\end{equation}
\begin{equation}
\boldsymbol{\epsilon}_n\approx \boldsymbol{\xi}_n + \boldsymbol{\mathfrak{I}}^{-1}_n\boldsymbol{\xi}
\label{eq:TTapprox}
\end{equation}
where $\boldsymbol{\mathfrak{I}}^{-1}$ in Equation (\ref{eq:TTapprox}) is Baker-Campbell-Hausdorff approximation \cite{barfoot2014} of Equation (\ref{eq:TT}) which is also referred to the left jacobian of $SE(3)$.

The pose fusion model above suggest{s} a framework for {the} $SE(3)$  pose fusion problem. 
{The multiple misalignment estimations in different places are fused by this framework. However instead of following the original batch fusion, we sequentially fuse posed until the uncertainty is small enough.}

Let $\textbf{T}_c,\boldsymbol{\Sigma}_c,\textbf{T}_k,\boldsymbol{\Sigma}_k$, respectively be $k_{th}$ new alignment estimation and fused alignment measurement up to the point {with their corresponding uncertainties}. Note that ideally $\textbf{T}_c$ and $\textbf{T}_k$ are identical as the new alignment estimation $\textbf{T}_k$ is always represented with respect to the first frame where the first alignment was estimated.  

Fusion of the two alignments can be found by applying {the} Gauss-Newton as 
\begin{equation}
\begin{split}
\textbf{A}& =\boldsymbol{\mathfrak{I}}^{-T}_c\boldsymbol{\Sigma}^{-1}_c\boldsymbol{\mathfrak{I}}^{-1}_c+\boldsymbol{\mathfrak{I}}^{-T}_k\boldsymbol{\Sigma}^{-1}_k\boldsymbol{\mathfrak{I}}^{-1}_k\\
\textbf{b}& =\boldsymbol{\mathfrak{I}}^{-T}_c\boldsymbol{\Sigma}^{-1}_c\boldsymbol{\xi}_c+\boldsymbol{\mathfrak{I}}^{-T}_k\boldsymbol{\Sigma}^{-1}_k\boldsymbol{\xi}_k\\
\boldsymbol{\xi}& =\textbf{A}^{-1}\textbf{b}.
\end{split}
\end{equation}
Then, the final fused misalignment $\textbf{T}^*$ is evaluated by iteratively updating $\textbf{T}^*$ with $\boldsymbol{\xi}$ as
\begin{equation}
\textbf{T}^* \leftarrow  \textbf{e}^{[\boldsymbol{\xi}]_{\times}}\textbf{T}^*
\end{equation}
where the updated alignment $\textbf{T}^*$ become{s} the new $\textbf{T}_c$ at the end of the iteration and utilized in the next fusion. The covariance of the next current alignment estimation is updated as follows.
\begin{equation}
\boldsymbol{\Sigma}_c = (\boldsymbol{\mathfrak{I}}^{-T}_c\boldsymbol{\Sigma}_c^{-1}\boldsymbol{\mathfrak{I}}^{-1}_c+\boldsymbol{\mathfrak{I}}^{-T}_k\boldsymbol{\Sigma}_k^{-1}\boldsymbol{\mathfrak{I}}^{-1}_k)^{-1}
\end{equation}

The fusion of the pose continues until $\boldsymbol{\Sigma}_c$ meet the predefined threshold.

\section*{Acknowledgment}

The authors gratefully acknowledge funding of the project by the CSIRO and QUT. The institutional support of CSIRO and QUT, and the help of several individual members of staff in the Robotics and Autonomous Systems Group, CSIRO, including Gavin Catt, Emili Hernandez, Ben Tam, Mark Cox, Tom Lowe and Alberto Elfes are greatly appreciated.

\ifCLASSOPTIONcaptionsoff
  \newpage
\fi

\bibliographystyle{IEEEtran}
\bibliography{ref}

\begin{IEEEbiography}[{\includegraphics[width=1in,height=1.25in,clip]{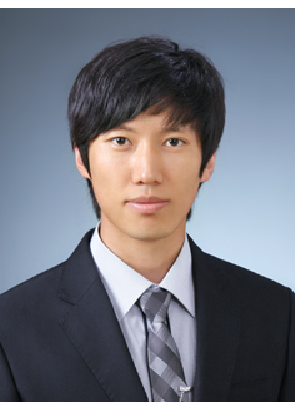}}]%
{Chanoh Park}
received the B.S. degree
in electrical engineering from Seoul
National University of Science and
Technology, Seoul, Korea, in 2010, and the M.S. degree in
computer science from the Sungkyunkwan University, Seoul, Korea, in 2012. He is currently
pursuing the Ph.D. degree in electrical engineering from
the Queensland University of Technology (QUT), Brisbane, QLD, Australia, and Commonwealth Scientific and Industrial Research Organisation (CSIRO), Pullenvale, QLD, Australia.
\end{IEEEbiography}

\begin{IEEEbiography}[{\includegraphics[width=1in,height=1.25in,clip]{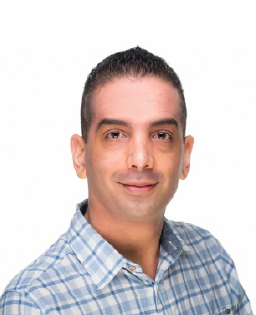}}]%
{Peyman Moghadam}
(Mâ07) received his Ph.D. degree in electrical and electronic engineering from Nanyang Technological University, Singapore. He is currently a Senior Research Scientist and Project Leader at the Robotics and Autonomous Systems Group, Commonwealth Scientific and Industrial Research Organisation (CSIRO), Australia. He is also an Adjunct Professor at the Queensland University of Technology (QUT). His current research interests include 3D multi-modal perception (3D++), SLAM, robotics, computer vision, and machine learning.
\end{IEEEbiography}

\begin{IEEEbiography}[{\includegraphics[width=1in,height=1.25in,clip]{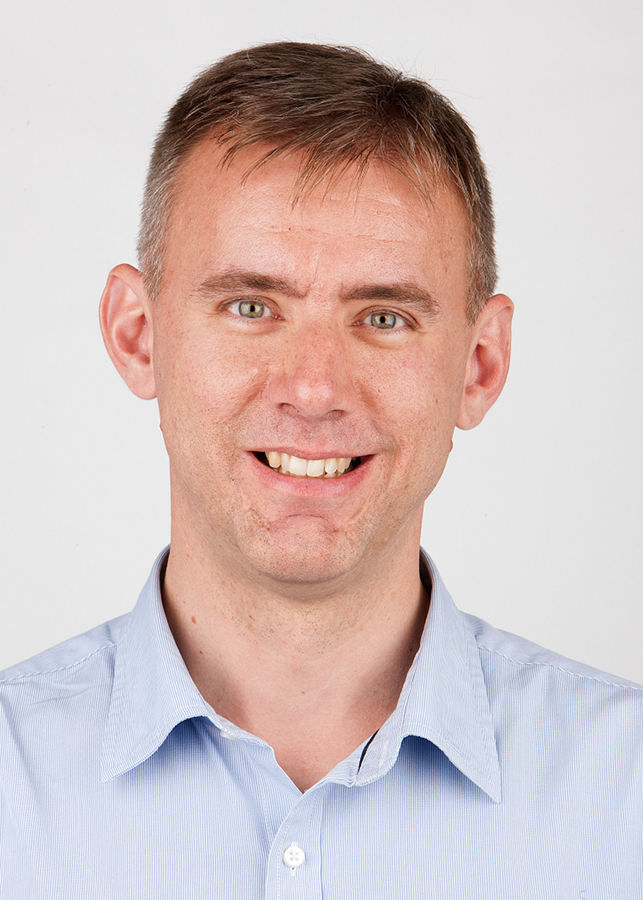}}]%
{Jason L. Williams}
(S01M07SM16) received degrees of BE(Electronics)/BInfTech from Queensland University of Technology, MSEE from the United States Air Force Institute of Technology, and PhD from Massachusetts Institute of Technology. He is currently a Senior Research Scientist in Robotic Perception at the Robotics and Autonomous Systems Group of Commonwealth Scientific and Industrial Research Organisation, Brisbane, Australia. His research interests include SLAM, computer vision, multiple object tracking and motion planning. He is also an adjunct associate professor at Queensland University of Technology. He previously worked in sensor fusion and resource management as a Senior Research Scientist at the Defence Science and Technology Group, Australia, and in electronic warfare as an engineering officer in the Royal Australian Air Force.
\end{IEEEbiography}

\begin{IEEEbiography}[{\includegraphics[width=1in,height=1.25in,clip,keepaspectratio]{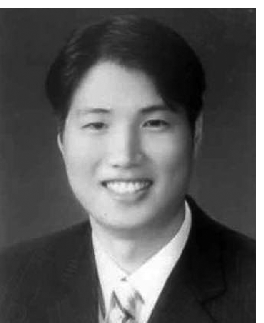}}]%
{Soohwan Kim}
 received a B.S. degree in Mechanical Engineering from Seoul National University in 2003, a M.S. degree in Computer Science from University of Southern California in 2004, and a Ph.D. degree in Engineering and Computer Science from The Australian National University in 2015. He was a Research Scientist in Electronics and Telecommunications Research Institute from 2005 to 2006, in Korea Science and Technology Institute from 2007 to 2011, in Bosch Research North America from 2015 to 2016, in CSIRO from 2016 to 2018, and a Senior Computer Vision Scientist in NVIDIA from 2018 to 2019. He is now an assistant professor in Sunmoon University. His research interests lie in the fields of robotics, computer vision, and machine learning.
\end{IEEEbiography}

\begin{IEEEbiography}[{\includegraphics[width=1in,height=1.25in,clip,keepaspectratio]{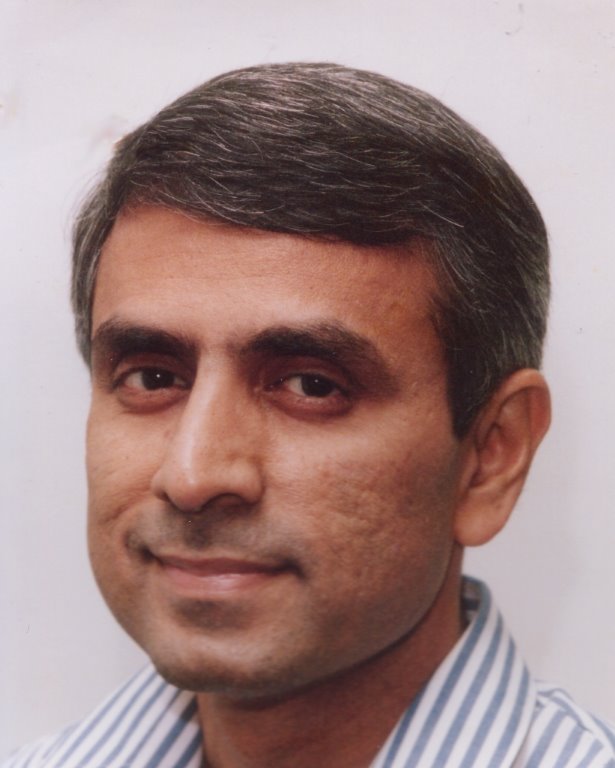}}]%
{Sridha Sridharan}
 (SMâ88) has a BSc (Electrical Engineering) degree and obtained a MSc (Communication Engineering) degree from the University of Manchester UK and a PhD degree in the area of Signal Processing from University of New South Wales, Australia. He is a Life Senior Member of the Institute of Electrical and Electronic Engineers - IEEE (USA).  He is currently with the Queensland University of Technology (QUT) where he is a Professor in the School Electrical Engineering and Robotics. Professor Sridharan is the Leader of the Research Program in Speech, Audio, Image and Video Technologies (SAIVT) at QUT. He has published over 500 papers consisting of publications in journals and in refereed international conferences in the areas of Image and Speech technologies. He has graduated over 80 PhD students at QUT. Prof Sridharan has also received a number of research grants from various funding bodies including commonwealth competitive funding schemes such as the Australian Research Council (ARC) and Cooperative Research Centres (CRC). Several of his research outcomes have been commercialised.   
\end{IEEEbiography}

\begin{IEEEbiography}[{\includegraphics[width=1in,height=1.25in,clip,keepaspectratio]{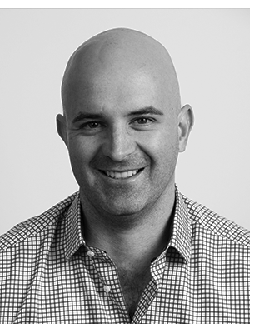}}]%
{Clinton Fookes}
(SMâ06) received his B.Eng. (Aerospace/Avionics), MBA, and Ph.D. degrees from the Queensland University of Technology (QUT), Australia. He is currently a Professor in Vision and Signal Processing within the Science and Engineering Faculty at QUT. He actively researchers across computer vision, machine learning, and pattern recognition areas. He serves on the editorial board for the IEEE Transactions on Information Forensics \& Security. He is a Senior Member of the IEEE, an Australian Institute of Policy and Science Young Tall Poppy, an Australian Museum Eureka Prize winner, and a Senior Fulbright Scholar. 
\end{IEEEbiography}

\end{document}